\documentclass[a4paper,11pt,numbers,sort&compress]{article}

\pdfoutput=1

\usepackage[paper=letterpaper,margin=1in]{geometry}
\usepackage{amsmath}
\usepackage{amssymb}
\usepackage{amsthm}

\usepackage{multirow}
\usepackage{cite,todonotes,ltablex,slashbox}

\usepackage{hyperref}
\hypersetup{pdftitle={hep-th}, 
      pdfauthor={Yang-Hui He, Vishnu Jejjala, Brent Nelson}, 
      pdfsubject={high energy theory, machine learning, linguistics},
      pdfstartview={FitH}, pdfpagelayout={TwoColumnRight},
      bookmarksopen, bookmarksnumbered, bookmarksopenlevel=2}

\def\IR{\mathbb{R}}

\def\cM{\mathcal{M}}

\def\hepth{\mbox{\sf{hep-th}}}
\def\hepph{\mbox{\sf{hep-ph}}}
\def\heplat{\mbox{\sf{hep-lat}}}
\def\mathph{\mbox{\sf{math-ph}}}
\def\grqc{\mbox{\sf{gr-qc}}}
\def\stat{\mbox{\sf{stat}}}
\def\qbio{\mbox{\sf{q-bio}}}
\def\qfin{\mbox{\sf{q-fin}}}
\def\condmat{\mbox{\sf{cond-mat}}}
\def\vhep{\mbox{\sf{viXra-hep}}}
\def\vqgst{\mbox{\sf{viXra-qgst}}}
\def\india{\mbox{\sf{Times-India}}}
\def\arXiv{\texttt{arXiv}}
\def\wordvec{\texttt{Word2vec}}

\def\tf{\mathop {\rm  tf}}
\def\idf{\mathop {\rm idf}}
\def\tfidf{\mathop {\rm tfidf}}

\newcommand{\w}[1]{\mbox{``#1''}}
\newcommand{\comment}[1]{}

\newcommand{\bn}[1]{\noindent{\color{blue}\textbf{BN:} #1}}

\newcommand{\yhh}[1]{\noindent{\color{green}\textbf{YHH:} #1}\\}

\theoremstyle{plain}
\newtheorem{definition}{Definition}
\newtheorem{theorem}{Theorem}

\begin{document}

\baselineskip=14pt
\parskip 5pt plus 1pt 

{}

\begin{center}        
  \LARGE\bf
 hep-th
\end{center}

\vspace{0.75cm}
\begin{center}        
	{\sc Yang-Hui He$^1$, Vishnu Jejjala$^2$, Brent D. Nelson$^3$} \\[4mm]
	\emph{\small 
	\begin{tabular}{rl}
	$^1$ &
		Department of Mathematics, City, University of London, EC1V 0HB, UK \\
		&Merton College, University of Oxford, OX14JD, UK\\
		&School of Physics, Nankai University, Tianjin,  300071, P.R.~China\\
	$^2$&
		Mandelstam Institute for Theoretical Physics, NITheP, CoE-MaSS, and\\
		& School of Physics, University of the Witwatersrand,\\
		& Johannesburg, WITS 2050, South Africa \\
	$^3$&
		Department of Physics, College of Science, Northeastern University, \\ 
		&Dana Research Center, 110 Forsyth Street, Boston, MA 02115, USA\\
	\end{tabular}	
	}
	\\[5mm]
	hey@maths.ox.ac.uk, \ vishnu@neo.phys.wits.ac.za, \ b.nelson@neu.edu
\end{center}

\vspace{1in}


\begin{abstract}
We apply techniques in natural language processing, computational linguistics, and machine-learning to investigate papers in hep-th and four related sections of the arXiv: hep-ph, hep-lat, gr-qc, and math-ph.
All of the titles of papers in each of these sections, from the inception of the arXiv until the end of 2017, are extracted and treated as a corpus which we use to train the neural network Word2Vec.
A comparative study of common $n$-grams, linear syntactical identities, word cloud and word similarities is carried out.
We find notable scientific and sociological differences between the fields. 
In conjunction with support vector machines, we also show that the syntactic structure of the titles in different subfields of high energy and mathematical physics are sufficiently different that a neural network can perform a binary classification of formal versus phenomenological sections with 87.1\% accuracy, and can perform a finer five-fold classification across all sections with 65.1\% accuracy.
\end{abstract}

\clearpage

\newpage

\tableofcontents

\section{Introduction and Summary}

The \arXiv~\cite{arxiv}, introduced by Paul Ginsparg to communicate preprints in high energy theoretical physics in 1991 and democratize science~\cite{pg}, has since expanded to encompass areas of physics, mathematics, computer science, quantitative biology, quantitative finance, statistics, electrical engineering and systems science, and economics and now hosts nearly $1.4$ million preprints.
In comparison with $123,523$ preprints archived and distributed in 2017~\cite{arxivstats}, around $2.4$ million scholarly articles across all fields of academic enquiry are published every year~\cite{paperstats}.
As a practitioner in science, keeping up with the literature in order to invent new knowledge from old is itself a formidable labour that technology may simplify.

In this paper, we apply the latest methods in computational linguistics, natural language processing and machine learning to preprints in high energy theoretical physics and mathematical physics to demonstrate important proofs of concept: by mapping words to vectors, algorithms can automatically sort preprints into their appropriate disciplines with over 65\% accuracy, and these vectors capture the relationships between scientific concepts.
In due course, many interesting properties of the word-vectors emerge and we make comparative studies across the different sub-fields.
Developing this technology will facilitate the use of computers as idea generating machines~\cite{denef}.
This is complementary to the role of computers in mathematics as proof assistants~\cite{voevodsky} and problem solvers~\cite{gg}.

The \textit{zeitgeist} of the moment has seen computers envelop our scientific lives.
In particular, in the last few years, we have seen an explosion of activity in applying artificial intelligence to various aspects of theoretical physics (cf.~\cite{ml} for a growing repository of papers).
In high energy theoretical physics, there have been attempts to understand the string landscape using deep neural networks~\cite{He:2017aed}, with genetic algorithms~\cite{Ruehle:2017mzq}, with network theory~\cite{Carifio:2017bov,Carifio:2017nyb}, for computing Calabi--Yau volumes~\cite{Krefl:2017yox}, for F-theory \cite{Wang:2018rkk}, and for CICY manifolds \cite{Bull:2018uow} etc.

Inspired by a recent effort by Evert van Nieuwenburg~\cite{EvN} to study the language of condensed matter theory, we are naturally led to wonder what the latest technology in language-processing using machine learning, utilized by the likes of \texttt{Google} and \texttt{Facebook}, would tell us about the language of theoretical physics.

As members of the high energy theoretical physics community, we focus on particular fields related to our own expertise: \hepth\ (high energy physics --- theory), \hepph\ (high energy physics --- phenomenology), \heplat\ (high energy physics --- lattice), \grqc\ (general relativity and quantum cosmology), and \mathph\ (mathematical physics).
In December 2017, we downloaded the titles and abstracts of the preprints thus far posted to \arXiv\ in these disciplines.
In total, we analyze a collection of some $395,000$ preprints.
Upon cleaning the data, we use \wordvec~\cite{word2vec1, word2vec2} to map the words that appear in this corpus of text to vectors in $\mathbb{R}^{400}$.
We then proceed to investigate this using the standard \texttt{Python} package {\sf gensim}~\cite{gensim}.
In parallel, a comparative study of the linguistic structure of the titles in the different fields in carried out.

It should be noted \footnote{
We thank Paul Ginsparg for kindly pointing out the relevant references and discussions.
} that there have been investigations of the \arXiv\ using textual analyses \cite{paul1,paul2,paul3}.
In this paper, we focus on the five sections of theory/phenomenology in the high-energy/mathematical physics community as well as their comparisons with titles outside of academia.
Moreover, we will focus on the syntactical identities which are generated from the contextual studies.
Finally, where possible we attempt to provide explanations why certain features in the data emerge, features which are indicative of the socio-scientific nature of the various sub-disciplines of the community.

We hope this paper will have readers in two widely-separated fields: our colleagues in the high energy theory community and those who study natural language processing in academia or in industry. As such, some guidance to the structure of what follows is warranted. For physicists, the introductory material in Section~\ref{sec:compute} will serve as a very brief summary and introduction to the vocabulary and techniques of computational textual analysis. Experts in this area can skip this entirely. Section~\ref{s:data} begins in Section~\ref{s:sets} with an introduction to the five \arXiv\ sections we will be studying. Our colleagues in physics will know this well and may skip this. Data scientists will want to understand our methods for preparing the data for analysis, which is described in Section~\ref{s:clean}, followed by some general descriptive properties of the \hepth\ vocabulary in Section~\ref{s:hepthfreq}. The analysis of \hepth\ using the results of \wordvec's neural network are presented in Section~\ref{sec:deep}, and those of the other four \arXiv\ sections in Section~\ref{sec:compare}. These two sections will be of greatest interest and amusement for authors who regularly contribute to the \arXiv, but we would direct data scientists to Section~\ref{sec:distance}, in which the peculiar geometrical properties of our vector space word embedding are studied, in a manner similar to that of the recent work by Mimno and Thompson~\cite{Mimno}. Our work culminates with a demonstration of the power of \wordvec, coupled with a support vector machine classifier, to accurately sort \arXiv\ titles into the appropriate sub-categories. We summarize our results, and speculate about future directions in Section~\ref{sec:conclude}.

\section{Computational Textual Analysis}
\label{sec:compute}

\subsection{Distributed Representation of Words}
We begin by reviewing some terminology and definitions (the reader is referred to~\cite{lecture,blogs} for more pedagogical material).
A {\bf dictionary} is a finite set whose elements are {\bf words}, each of which is an ordered finite sequence of letters.
An {\bf $n$-gram} is an ordered sequence of $n$ words.
A sentence can be considered an $n$-gram.
Moreover, since we will not worry about punctuation, we shall use these two concepts interchangeably.	
Continuing in a familiar manner, an ordered collection of sentences is a {\bf document} and a collection (ordered if necessary) of documents is a {\bf corpus}:
\begin{equation}
\mbox{word} \in \mbox{$n$-gram} \subset \mbox{document} \subset \mbox{corpus}  \ .
\end{equation}
Note that for our purposes the smallest unit is ``word'' rather than ``letters'' and we will also not make many distinctions among the three set inclusions, \textit{i.e.}, after all, we can string the entire corpus of documents into a single $n$-gram for some very large $n$.
As we shall be studying abstracts and titles of scientific papers, $n$ will typically be no more than $\mathcal{O}(10)$ for titles, and $\mathcal{O}(100)$ for abstracts.

In order to perform any analysis, one needs a numerical representation of words and $n$-grams;
this representation is often called a {\bf word embedding}.
Suppose we have a dictionary of $N$ words.
We can lexicographically order them, for instance, giving us a natural vector of length $N$.
Each word is a vector containing precisely a single $1$, corresponding to its position in the dictionary and $0$ everywhere else.
The entire dictionary is thus the $N \times N$ identity matrix and the $i$-th word is simply the elementary basis vector $e_i$.
This representation is sometimes called the {\bf one-hot} form of a vector representation of words (the hot being the single $1$-entry).
This representation of words is not particularly powerful because not much more than equality-testing can be performed.

The insight of \cite{word2vec} is to have a weighted vectorial representation $\vec{v_w}$, constructed so as to reflect the {\it meaning} of the word $w$.
For instance, $\vec{v}_{\w{car}} - \vec{v}_{\w{cars}}$ should give a close value to $\vec{v}_{\w{apple}} - \vec{v}_{\w{apples}}$.
Note that, for the sake of brevity, we will be rather lax about the notation on words: \textit{i.e.}, $\vec{v}_{\w{word}}$ and ``word'' will be used interchangeably.
Indeed, we wish to take advantage of the algebraic structure of a vector space over $\IR$ to allow for addition and subtraction of vectors.
An archetypal example in the literature is that 
\begin{equation}\label{king}
\w{king} - \w{man} + \w{woman} = \w{queen} \ .
\end{equation}
In order to arrive at a result such as \eqref{king}, we need a few non-trivial definitions.
\begin{definition}
A {\bf context window} with respect to a word $W$, or specifically, a \textit{$k$-around context window} with respect to $W$, is a subsequence within a sentence or $n$-gram containing the word $W$, containing all words a distance $k$ away from $W$, in both directions. Similarly, a $k$ context window, without reference to a specific word, is simply a subsequence of length $k$.
\end{definition}
For example, consider the following $46$-gram, taken from the abstract of~\cite{Witten:1991zd}, which is one of the first papers to appear on \hepth\ in 1991:
\begin{quote}
{\sf
String theories with two dimensional space-time target spaces are characterized by the existence of a ground ring of operators of spin (0,0). By understanding this ring, one can understand the symmetries of the theory and illuminate the relation of the critical string theory to matrix models.}
\end{quote}
In this case, a $2$-around  context window for the word ``spaces'' is the 5-gram ``space-time target spaces are characterized''.
Meanwhile, $2$-context windows are $2$-grams such as ``ground ring'' or ``(0,0)''.
Note that we have considered hyphenated words as a single word, and that the $n$-gram has crossed between two sentences.

The above example immediately reveals one distinction of scientific and mathematical writing that does not often appear in other forms of writing: the presence of mathematical symbols and expressions in otherwise ordinary prose. Our procedure will be to ignore all punctuation marks in the titles and abstracts. Thus, ``(0,0)'' becomes the `word' ``00''. Fortunately, such mathematical symbols are generally rare in the abstracts of papers (though, of course, they are quite common the body of the works themselves), and even more uncommon in the titles of papers.


The correlation between words in the sense of context, reflecting their likely proximity in a document, can be captured by a matrix:
\begin{definition}
The {\bf $k$ co-occurrence matrix} for an $n$-gram is a symmetric, $0$-diagonal square matrix (of dimension equalling the number of unique words) which encodes the adjacency of the words within a $k$ context window. 
\end{definition}
\noindent Continuing with the above example of the $46$-gram, we list all the unique words horizontally and vertically, giving a $37 \times 37$ matrix for this sentence (of course, one could consider the entire corpus as a single $n$-gram and construct a much larger matrix).
The words can be ordered lexicographically, \{``a'', ``and'', ``are'', \ldots \}.
One can check that in a $2$-context-window (testing whether two words are immediate neighbors), for instance, ``a'' and ``and'' are never adjacent; thus, the $(i,j) = (1,2)$ entry of the $2$ co-occurrence matrix is $0$.

\subsection{Neural Networks}
The subject of neural networks is vast and into its full introduction we certainly shall not delve.
Since this section is aimed primarily at theoretical physicists and mathematicians, it is expedient only to remind the reader of the most rudimentary definitions, enough to prepare for the ensuing subsection.
We recall first that:
\begin{definition}
A {\bf neuron} is a (typically analytic) function $f(\sum_{i=1}^k w_i \cdot x_i +b)$ whose argument $x_i$ is the {\it input}, typically some real tensor with multi-index $i$.
The value of $f$ is called the {\it output}, the parameter $w_i$ is some real tensor called {\it weights} with $\cdot$ appropriate contraction of the indices and the parameter $b \in \IR$ is called the {\it off-set} or {\it bias}. 
\end{definition}
The function $f$ is called an activation function and often takes the non-linear forms of a hyperbolic tangent or a sigmoid.
The parametres $w_i$ and $b$ are to be determined empirically as follows:
\begin{itemize}
\item Let there be a set of input values $x_i^{j}$ labelled by $j = 1, \ldots, T$ such that the output $y^{j}$ is known: $\{x_i^{j}, y^{j}\}$ is called the {\bf training set};
\item Define an appropriate measure of goodness-of-fit, typically one uses the standard deviation (mean-square-error)\\
\[
Z(w_i, b) := \sum\limits_{j=1}^T \big( f(\sum_{i=1}^k w_i \cdot x_i^j +b) - y^j \big)^2 \ ;
\]
\item Minimize $Z(w_i, b)$ as a function of the parameters (often numerically by steepest descent methods), whereby determining the parameters 
$(w_i, b)$;
\item Test the now-fixed neuron on unseen input data.
\end{itemize}
At this level, we are doing no more than (non-linear) regression by best-fit.
We remark that in the above definition, we have made $f$ a real-valued function for simplicity. In general, $f$ will be itself tensor-valued.

The power of the neural network comes from establishing a network structure of neurons --- much like the complexity of the brain comes from the myriad inter-connectivities among biological neurons:
\begin{definition}
A {\bf neural network}  is a finite directed graph, with possible cycles, each node of which is a neuron, such that the input for the neuron at tail of an arrow is the output of the neuron at the head of the arrow.
We organize the neural network into {\em layers} so that
\begin{enumerate}
\item The collection of nodes with arguments explicitly involving the actual input data is called the {\it input layer};
\item Likewise, the collection of nodes with arguments explicitly involving the actual output of data is called the {\it output layer};
\item The collection of all other nodes are called {\it hidden layers}.
\end{enumerate}
\end{definition}
Training the neural network proceeds as the algorithm above, with the input and output layers interacting with the training data and each neuron giving its own set of weight/off-set parameters.
The measure $Z$ will thus be a function in many variables over which we optimize.

One might imagine the design of a neural network -- with its many possible hidden layers, neuron types, and internal architecture -- would be a complicated affair, and highly-dependent on the problem at hand. But in this we can take advantage of the powerful theorem by Cybenko and Hornik~\cite{approx}:
\begin{theorem} {\bf Universal Approximation Theorem} [Cybenko-Hornik]
A neural network with a single hidden layer with a finite number of neurons can approximate any neural network in the sense of prescribing a dense set within the set of continuous functions on $\IR^n$.
\end{theorem}

\subsection{Word2Vec}\label{s:word2vec}

We now combine the concepts in the previous two subsections to use a neural network to establish a {\it predictive} embedding of words. In particular, we will be dealing with {\sf Word2Vec}~\cite{word2vec}, which has emerged as one of the most sophisticated models for natural language processing (NLP). The {\sf Word2Vec} software can utilize one of two related neural network models:  (1) the {\bf continuous bag of words [CBOW]} model, or (2) the {\bf skip-gram [SG]} model, both consisting of only a single hidden layer, which we will shortly define below in detail.

The CBOW approach is perhaps the most straight-forward: given an $n$-gram, form the multiset (`bag') of all words found in the $n$-gram, retaining multiplicity information. One might also consider including the set of all $m$-grams ($m\leq n$) in the multiset. A collection of such `bags', all associated with a certain classifier (say, `titles of \hepth\ papers'), then becomes the input upon which the neural network trains.

The SG model attempts to retain contextual information by utilizing context windows in the form of {\bf $k$-skip-$n$-grams}, which are $n$ grams in which each of the components of the $n$-gram occur at a distance $k$ from one another. That is, length-$k$ gaps exist in the $n$-grams extracted from a piece of text. The fundamental object of study is thus a  $k$-skip-$n$-gram, together with the set of elided words. It is the collection of these objects that becomes the input upon which the neural network trains.

The two neural network approaches are complementary to one another. Coarsely speaking, CBOW is trained to predict a word given context, while SG is trained to predict context given a word. To be more concrete, a CBOW model develops a vector space of word embeddings in such a way as to maximize the likelihood that, given a collection of words, it will return a single word that best matches the context of the collection. By contrast, the vector space of word embeddings constructed by the SG model is best suited to give a collection of words likely to be found in context with a particular word.

While CBOW and SG have specific roles, the default method for generating word embeddings in a large corpus is to use the CBOW model. This is because it is clearly more deterministic to have more input than output, \textit{i.e.}, a single-valued function is easier to handle than a multi-valued one. This will be our choice going forward. Note that our ultimate interest is in the information contained in the word embedding itself, and not in the ability of the model to make predictions on particular words. When we attack the document classification problem via machine learning, in Section~\ref{sec:class}, we will use the distributed representation generated by {\sf Word2Vec} as training data for a simple support vector machine classifier.

\begin{figure}[t]
\centerline{
\includegraphics[trim=0mm 0mm 0mm 0mm, clip, width=5in]{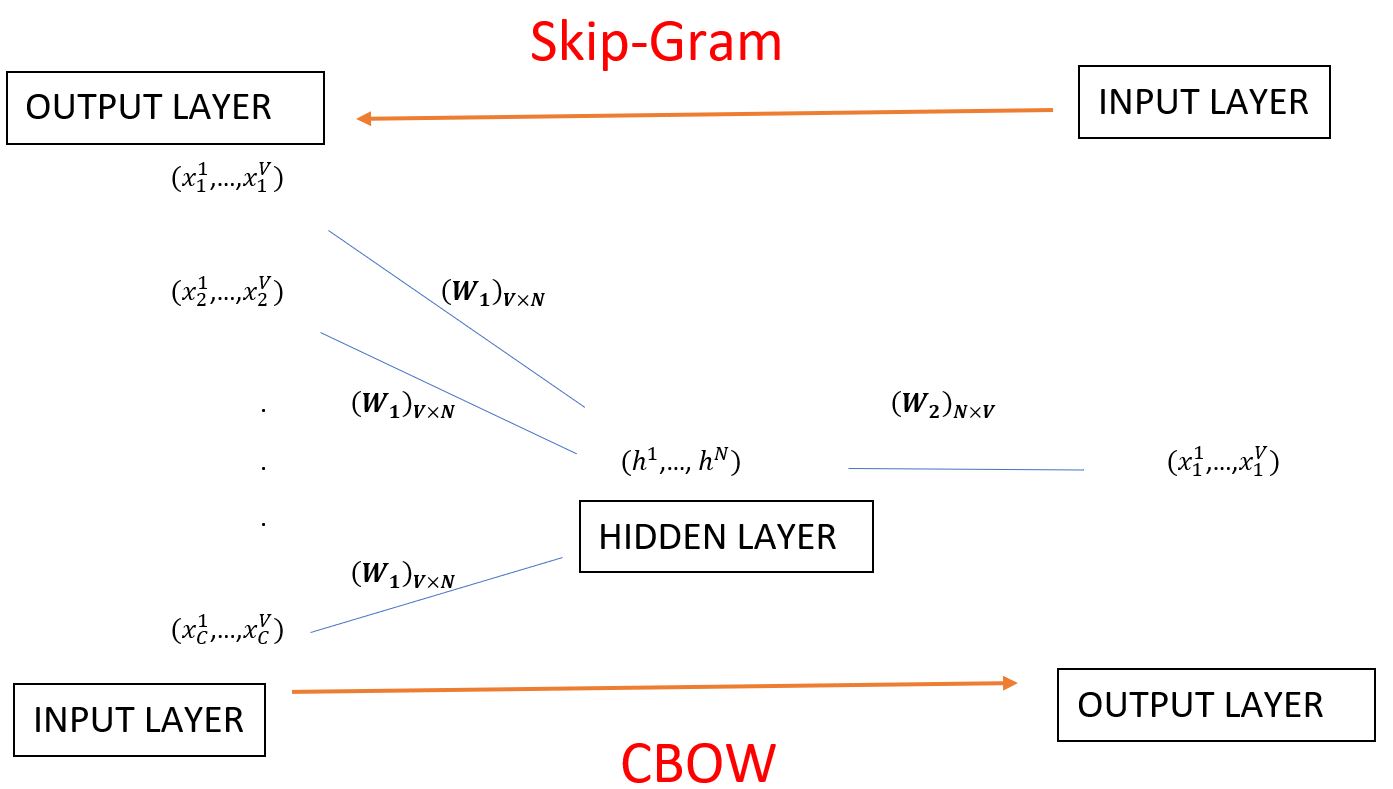}
}
\caption{{\sf {\small
The neural network consists of a single hidden layer, which constructs a mapping from $\IR^V$ to $\IR^V$. The parameters to be fit by the neural network are the transformation matrices $W_1$ and $W_2$, with the entries of $W_1$ constituting the word embedding into the space $\IR^N$. After training, a CBOW model takes multiple $\vec{x}$ inputs, associated with a particular context $C$, and maps them to a single output vector. Conversely, after training, a SG model takes a single word $\vec{x}$ as an input, and returns the set of vectors $\vec{x}_C$ associated with the appropriate context.
\label{f:cbowsg}}}}
\end{figure}

With the CBOW model in mind, therefore, let us consider the structure of the underlying neural network, depicted schematically in~Figure \ref{f:cbowsg}.
Let $V$ denote the vocabulary size so that any word can be \textit{ab initio} trivially represented by a one-hot vector in $\IR^V$. The dimension $V$ could be rather large, and we will see how a reduction to an $N$-dimensional representation is achieved. The overall structure has three layers:
\begin{enumerate}
\item The input layer is a list of $C$ context words. Thus, this is a list of $C$ vectors 
	$\vec{x}_{c=1,\ldots,C}$ each of dimension $V$. We denote these component-wise as $x_{c=1,\ldots,C}^{i=1,\ldots,V}$. This list is the ``bag of words'' of CBOW, and in our case each such list will represent a single, contextually-closed object, such as a paper title or paper abstract.
\item The output layer is a single vector $\vec{y}$ of dimension $V$. We will train the neural network with a large number of examples where $y$ is known, given the words $\vec{x}$ within a context $C$, so that one could thence {\it predict} the output. Continuing with our example of~\cite{Witten:1991zd}, the title of this paper provides a specific context $C$: ``Ground ring of two-dimensional string theory''. Thus for an input of \{``ring'', ``of'', ``string'', ``theory''\}, we ideally wish to return the vector associated with ``two-dimensional'' when the window size is set to two. Of course, we expect words like ``theory'' to appear in many other titles. The neural network will therefore optimize over many such titles (contexts) to give the ``best'' vector representation of words.
\item There is a single hidden layer consisting of $N$ neural nodes. The function from the input layer to the hidden layer is taken to be linear map, \textit{i.e.}, a $V \times N$ matrix $W_1$ of weights. Likewise, the map from the hidden layer to the output layer is an $N \times V$ weight matrix $W_2$. Thus, in total, we have $2V\times N$ weights which will be optimized by training. Note that $N$ is a fixed parameter (or hyper-parameter) and is a choice. Typically, $N \sim 300 - 500$ has been shown to be a good range~\cite{lecture,blogs,course}; in this paper we take $N = 400$.
\end{enumerate}

To find the optimal word embedding, or most faithful representation of the words in $\IR^N$, each input vector $\vec{x}_c$, in a particular context (or bag), is mapped by $W_1$ to an $N$ vector (the actual word-vector after the neural network is trained) $\vec{h} =  [\vec{x}_c]^T \cdot W_1$. 
Of course, because $\vec{x}$ is a Kronecker-delta, $\vec{h}$ is just the $k$-th row of $W_1$ where $k$ is the only component equal to $1$ in $\vec{x}_c$. A measure of the proximity between an input and output word-vector is the weighted inner product
\begin{equation} 
\left< \vec{x}_c, \ \vec{y} \right> := [\vec{x}_c]^T \cdot W_1 \cdot W_2 \cdot \vec{y}\, .
\end{equation} 
Hence, for each given context $C_\alpha$, where $\alpha$ might label contextually-distinct objects (such as paper titles) in our training set, we can define a score for each component $i$ (thus in the one-hot representation, each word) $u_c^{j = 1, \ldots, V}$ in the vocabulary as
\begin{equation} u^j_c := [\vec{x}_c]^T \cdot W_1 \cdot W_2\, .
\end{equation}
As always with a list, one can convert this to a probability (for each $c$ and each component $j$) via the {\it softmax function}:
\begin{equation}
		p(u^j_c | x_c) := \frac{\exp(u^j_c)}{\sum\limits_{j=1}^V \exp(u^j_c)}\, .
\end{equation}

Finally, the components of the output vector  $\vec{y}$ is the product over the context words of these probabilities
\begin{equation}
	y^j = \prod\limits_{c=1}^C \frac{\exp(u^j_c)}{\sum\limits_{j=1}^V \exp(u^j_c)} \, .
\end{equation}
The neural network is trained by maximizing the log-likelihood of the probabilities across all of our training contexts
\begin{equation}\label{Zcbow}
Z(W_1, W_2) := \frac{1}{|D|} \sum\limits_{\alpha=1}^{|D|} \log \prod\limits_{c=1}^{C_{\alpha}}
	\frac{\exp([\vec{x}_c]^T \cdot W_1 \cdot W_2)}{\sum\limits_{j=1}^V \exp([\vec{x}_c]^T \cdot W_1 \cdot W_2)}\, ,
\end{equation}
where we have written the functional dependence in terms of the $2V\times N$ variables of $W_1$ and $W_2$ because we need to extremize over these. In~(\ref{Zcbow}), the symbol $|D|$ represents the number of independent contexts in the training set (i.e., $\alpha = 1,\dots,|D|$).

We will perform a vector embedding study as discussed above, and perform various contextual analyses using the bag-of-words model. Fortunately, many of the the requisite algorithms have been implemented into {\sf python} with the {\sf gensim} package~\cite{gensim}.
%


\subsection{Distance Measures}
Once we have represented all words in a corpus as vectors in $\IR^N$, we will loosely use the ``$=$'' sign to denote that two words are ``close'' in the sense that the Euclidean distance between the two vectors is small. In practice, this is measured by computing the cosine of the angle between the vectors representing the words. That is, given words $w_1$ and $w_2$, and their associated word vectors $\vec{v}_{w_1}$ and $\vec{v}_{w_2}$, we can define distance as
\begin{equation}
d(w_1,w_2) := \frac{\vec{v}_{w_1} \cdot \vec{v}_{w_2}}{\left|\vec{v}_{w_1}\right| \left|\vec{v}_{w_2} \right|} \ .
\label{distance}
\end{equation}
Generically, we expect $d(w_1,w_2)$ will be close to zero, meaning that two words are not related. However, if $d(w_1,w_2)$ is close to $+1$, the words are similar. Indeed, for the same word $w$, tautologically $d(w,w) = 1$. If $d(w_1,w_2)$ is close to $-1$,  then the words are dissimilar in the sense that they tend to be far apart in any context window. We will adopt, for clarity, the following notation:
\begin{definition}\label{sim}
Two words $w_1$ and $w_2$ are 
\begin{description}
\item[similar] in the sense of $d(w_1,w_2) \sim 1$ (including the trivial case of equality), and are denoted as $w_1 =  w_2$; and
\item[dissimilar] in the sense of $d(w_1,w_2) \sim -1$, and are denoted as $w_1 \neq w_2$.
\end{description}
Vector addition generates signed relations such as $w_1 + w_2 = w_3$, $w_1 + w_2 - w_3 = w_4$, etc.
We will call these relations {\bf linear syntactic identities}.
\end{definition}
For instance, our earlier example of \eqref{king} is one such identity involving four words.
Henceforth, we will bear in mind that ``='' denotes the {\it closest} word within context windows inside the corpus.
Finding such identities in the \hepth\ \arXiv\ and its sister repositories will be one of the goals of our investigation.

As a technical point, it should be noted that word-vectors {\it do not span a vector space} $V$ in the proper sense.
First, there is no real sense of closure, one cannot guarantee the sum of vectors is still in $V$, only the closest to it by distance. Second, there is no sense of scaling by elements of the ground field, here $\IR$. In other words, though the components of word-vectors are real numbers,
it is not clear what $a w_1 + b w_2$ for arbitrary $a,b \in \IR$ means.
The only operation we can perform is the one discussed above by adding two and subtracting two vectors in the sense of a syntactic identity.

\subsection{Term Frequency and Document Frequency}
\label{sec:tfidf}

Following our model of treating the set of titles/abstracts of each section as a single document, one could thus consider the \arXiv\ as a corpus. The standard method of cross-documentary comparisons is the so-called {\bf term-frequency - inverse document frequency (tf-idf)}, which attempts to quantify the importance of a particular word in the corpus overall:
\begin{definition}
Let $D$ be a corpus of documents, $d \in D$ a document and $t \in d$ be a word (sometimes also called a term) in $d$. Let $f(t,d) := |x \in d : x = t|$ represent the raw count of the number of appearances of the word $t$ in the document $d$, where the notation $|X|$ means the cardinality of the set represented by $X$.
\begin{itemize}
\item The {\bf term frequency} $\tf(t,d)$ is a choice of function of the count of occurrences $f(t,d) := |x \in d : x = t|$ of $t$ in $d$;
\item The {\bf inverse document frequency} $\idf(t,D)$ is the minus logarithm of the fraction of documents containing $t$:
\[
\idf(t,D) := -\log \frac{|d \in D : t \in d|}{|D|} = \log \frac{|D|}{|d \in D : t \in d|} \ ;
\]
\item The {\bf tf-idf} is the product of the above two: $\tfidf(t,d,D) := \tf(t,d) \cdot \idf(t,D)$ \ .
\end{itemize}
\end{definition}
\noindent In the context of our discussion in Section~\ref{s:word2vec}, $D$ might represent the sum of all titles, and each $d$ might represent a distinct title (or ``context'').

The simplest $\tf$ is, of course, to just take $f(t,d)$ itself. Another commonly employed $\tf$ is a logarithmically scaled frequency $\tf(t,d) = \log(1 + f(t,d))$, which we shall utilize in our analysis. Thus we will have
\begin{equation}
\tfidf(t,d,D)  =\log( 1 + |x \in d : x = t| )  \log \frac{|D|}{|d \in D : t \in d|} \in \IR_{\geq 0}\, ,
\end{equation}
where $d$ might represent the string of words in all titles in a given \arXiv\ section, labeled by $D$.
The concept of weighting by the inverse document frequency is to penalize those words which appear in virtually all documents. Thus a tf-idf score of zero means that the word either does not appear at all in a document, or it appears in {\it all} documents.



\section{Data Preparation}
\label{s:data}

\subsection{Data Sets}
\label{s:sets}

As mentioned in the introduction, we will be concerned primarily with the language of \hepth, but we will be comparing this section of the \arXiv\ with closely-related sections. The five categories that will be of greatest interest to us will be:
\begin{description}
\item[\hepth] Begun in the summer of 1991, \hepth\ was the original preprint listserv for theoretical physics. Traditionally the content has focused on formal theory, including (but not limited to), formal results in supersymmetric field theory, string theory and string model building, and conformal field theory.
\item[\hepph] Established in March of 1992, \hepph\ was the bulletin board designed to host papers in phenomenology -- a term used in high energy theory to refer to model-building, constraining known models with experimental data, and theoretical simulation of current or planned particle physics experiments.
\item[\heplat] Launched in April of 1992, \heplat\ is the \arXiv\ section dedicated to numerical calculations of quantum field theory observables using a discretization of space-time (``lattice'') that allows for direct computation of correlation functions that cannot be computed by standard perturbative (Feynman diagram) techniques. While closely related to the topics studied by authors in \hepth\ and \hepph, scientists posting research to \heplat\ tend to be highly-specialized, and tend to utilize high performance computing to perform their calculations at a level that is uncommon in the other \arXiv\ sections.
\item[\grqc] The bulletin board for general relativity and quantum cosmology, \grqc, was established in July of 1992. Publications submitted to this section of \arXiv\ tend to involve topics such as black hole solutions to general relativity in various dimensions, treatment of spacetime singularities, information theory in general relativity, and early universe physics. Explicit models of inflation, and their experimental consequences, may appear in \grqc, as well as in \hepth\ and (to a lesser extent) \hepph.
Preprints exploring non-string theory approaches to quantum gravity typically appear here.
\item[\mathph] The youngest of the sections we consider, \mathph\ was born in March of 1998 by re-purposing a section of the general physics \arXiv\ then called ``Mathematical Methods in Physics''. As the original notification email advised, ``If you are not sure whether or not your submission is physics, then it should be sent to math-ph.'' Today, papers submitted to \mathph\ are typically the work of mathematicians, whose intended audience is other mathematicians, but the content of which tends to be of relevance to certain areas of string theory and formal gauge theory research.
\end{description}

Given that part of the motivation for the current work is to use language as a marker for understanding the social dynamics -- and overlapping interests -- of our community, it is relevant to mention a few important facts about the evolution of the \arXiv. The \arXiv\ repository began originally as a listserv, which sent daily lists of titles and abstracts of preprints to its subscribers. The more technically savvy of these recipients could then seek to obtain these preprints via anonymous \mbox{\sf{ftp}} or \mbox{\sf{gopher}}. A true web-interface arrived at the end of 1993.

\begin{figure}[t]
\centerline{
\includegraphics[trim=15mm 0mm 0mm 0mm, clip, width=7in]{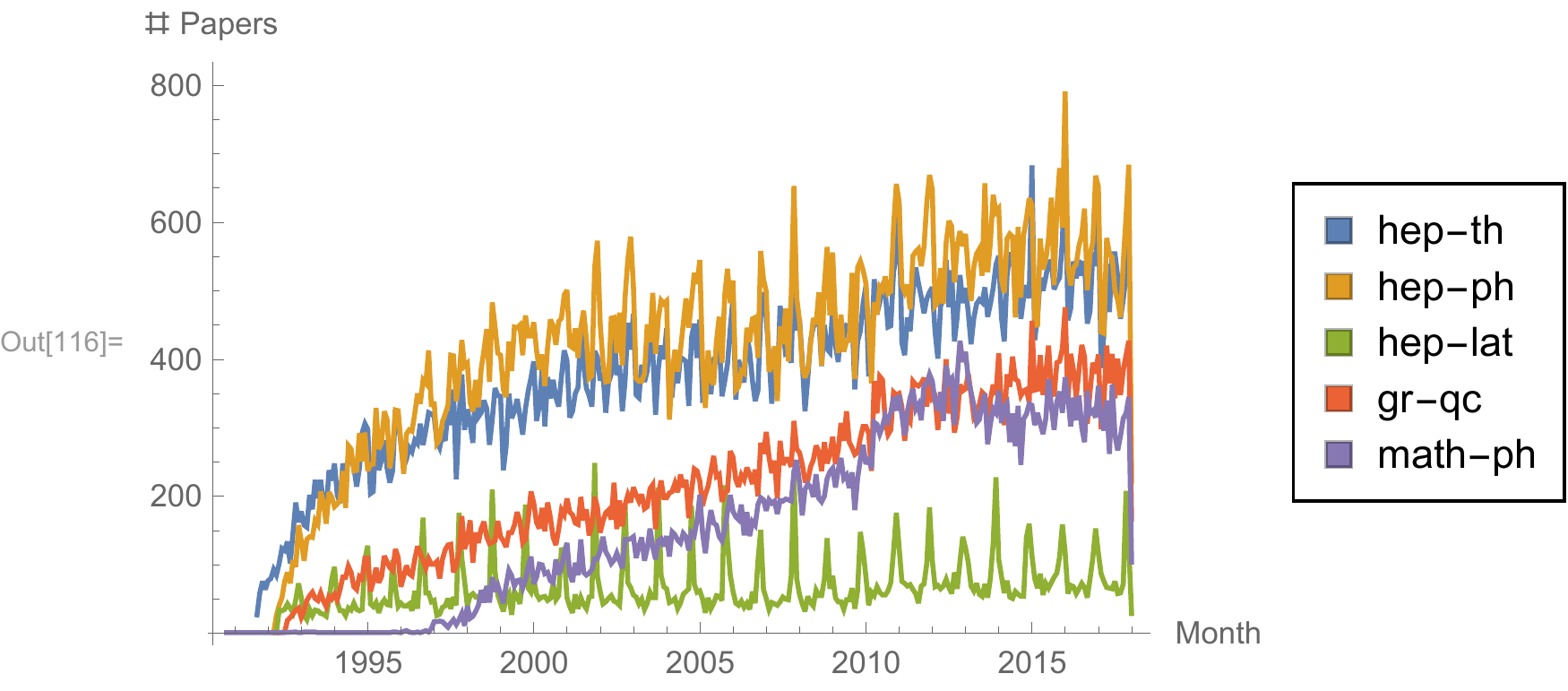}
}
\caption{{\sf {\small
Number of papers published for five related sections of high energy physics, since the beginning of the \arXiv\ in 1991, until the end of December, 2017. 
\label{f:number}}}}
\end{figure}


New sections of the \arXiv\ proliferated rapidly in the early 90's, at the request of practicing physicists. Given the limited bandwith -- both literal and metaphorical -- of most university professors, it seemed optimal to sub-divide the \arXiv\ into ever smaller and more focused units. Thus, research was pigeonholed into ``silos'' by design. As a result, individual faculty often came to identify strongly with the section of \arXiv\ to which they regularly posted. While cross-listing from a primary section to a secondary (and even a tertiary) section began in May of 1992, such cross-listing was generally rare throughout much of the early years of the \arXiv. The total number of publications per month, in these five sections of \arXiv, is shown in Figure~\ref{f:number}. The reader is also referred to \arXiv\ itself for a detailed analysis of such statistics.


\begin{table}[t]
\centering
\begin{tabular}{| c | c || c | c || c | c |}
\hline \arXiv & No. of & \multicolumn{2}{|c||}{Titles}  & \multicolumn{2}{|c|}{Abstracts}  \\
Section & Papers & Mean Length & Unique Words & Mean Length & Unique Words \\
\hline \hline
\hepth & 120,249 & 8.29 & 37,550 & 111.2 & 276,361 \\
\hepph & 133,346 & 9.34 & 46,011 & 113.4 & 349,859 \\
\heplat & 21,123 & 9.31 & 10,639 & 105.7 & 78,175 \\
\grqc & 69,386 & 8.74 & 26,222 & 124.4 & 194,368 \\
\mathph & 51,747 & 9.19 & 28,559 & 106.1 & 194,149 \\
\hline
\end{tabular}
\caption{{\sf {\small Gross properties of the five \arXiv\ sections studied in this paper. Numbers includes all papers through the end of 2017. The count of unique words does not distinguish upper and lower-case forms of a word. ``Length'' here means the number of words in a given title or in a given sentence in the abstract. 
As anticipated, there is remarkable similarity across these five sections of \arXiv\ for the mean lengths.}}}
\label{tab:wordcount}
\end{table} 


We extracted metadata from the \arXiv\ website, in the form of titles and abstracts for all submissions, using techniques described in~\cite{EvN}. The number of publications represented by this dataset is given in Table~\ref{tab:wordcount}. Also given is the mean number of words in the typical title and abstract of publications in each of the five sections, as well as the count of unique words in each of the sections. 

Titles and abstracts are, of course, different categories serving different functions. To take a very obvious example, titles do not necessarily need to obey the grammatical rules which govern standard prose. Nevertheless, we can consider each title in \hepth, or any other section of the \arXiv, as a sentence. For \hepth\ this gives us the {\bf raw data} of $120,249$ sentences.

Abstracts are inherently different. They represent a collection of sentences which cluster around particular semantic content. Grammatically, they are quite different from the titles. As an example, being comprised of full sentences suggests that punctuation is meaningful in the abstract, while generally irrelevant in titles. Finally, whereas each of the 120,249 titles in \hepth\ can be though of as semantically distinct sentences in a single document (the entirety of \hepth), the abstracts must be thought of as individual documents within a larger corpus. This notion of ``grouping'' sentences into semantic units can make the word embedding process more difficult.

While variants of {\bf Word2Vec} exist which can take this nuance into account, we will simply aggregate {\it all} abstracts on each section of the \arXiv, then separate them only by full sentences. For \hepth, the abstracts produce 608,000 sentences over all 120,249 papers, comprising 13,347,051 words, of which 276,361 are unique. 

\subsection{Data Cleaning: Raw, Processed and Cleaned}
\label{s:clean}

As any practicing data scientist will attest, cleaning and pre-processing raw data is a crucial step in any analysis in which machine learning is to be utilized. The current data set is no exception. Indeed, some data preparation issues in this paper are likely to be unique in the natural language processing literature. In this subsection we describe the steps we took to prepare the data for neural network analysis.

Our procedure for pre-processing data proceeded along the following order of operations:
\begin{enumerate}
\item Put everything into lower case;
\item Convert all key words, typically nouns, to singular case. Indeed, it typically does not make sense to consider the words
	``\verb|equation|'' and ``\verb|equations|'' as different concepts;
\item Spellings of non-English names, including \LaTeX\, commands, are converted to standard English spelling.
	For example, ``\verb|schroedinger|'', ``\verb|schr\"odinger|'' and ``\verb|schr"odinger|'' will all be replaced by simply
	``\verb|schrodinger|''  (note that at this stage we already do not have any further upper case letters so the ``s'' is not capitalized);
\item At this stage, we can replace punctuation such as periods, commas, colons, etc, as well as \LaTeX\ backslash commands such as \verb|\c{}| which do appear (though not often) in borrowed words such as ``aper\c{c}u'', as well as \verb|\cal| for calligraphic symbols (which do appear rather often), such as in ``${\cal N} = 4$ susy''. Note that we keep parentheses intact because words such as \verb|su(n)| appear often; so too we will keep such \LaTeX\ commands as \verb|^| and \verb|_| because superscripts and subscripts, when they appear in a title, are significant;
\item Now, we reach a highly non-trivial part of the replacements: including important technical acronyms. Though rarely used directly in titles, acronyms are common in our field. All acronyms serve the purpose of converting an $n$-gram into a single monogram. So, for example, ``\verb|quantum field theory|'' should appear together as a single unit, to be replaced by ``\verb|qft|''. This is a special case of bi-gram and tri-gram grouping, to be discussed below. In other cases, shorthand notation allows for a certain blurring between subject and adjective forms of a word. Thus,``\verb|supersymmetry|'' and ``\verb|supersymmetric|'' will become ``\verb|susy|''.
Note that such synonym studies were carried out in \cite{paulstudent}.
\end{enumerate}

At this stage, we use the term {\bf processed data} to refer to the set of words. One could now construct a meaningful word embedding, and use that embedding to study many interesting properties of the dataset. However, for some purposes it is useful to do further preparation, so as to address the aspects of \arXiv\ that are particularly scientific in nature. We thus performed two further stages of preparation on the data sets of paper titles, only.

First we remove any conjunctions, definite articles, indefinite articles, common prepositions, conjugations of the verb ``to be'' etc., because they add no scientific content to the titles. We note, however, that one could argue that they add some grammatical content and could constitute a separate linguistic study. Indeed, we will restore such words as part of our analysis in Section~\ref{sec:compare}.

In step \#5 above, certain words were manually replaced with acronyms commonly used in the high energy physics community. However, there are certain are bi-grams and tri-grams that -- while sometimes shortened to acronyms -- are clearly intended to represent a single concept.
One can clearly see the advantage of merging certain word pairs into compound mono-grams for the sake of textual analysis. For example, one would never expect to see the word ``Carlo'' in a title which was not preceded by the word ``Monte''. Indeed, the vast majority of $n$-grams involving proper nouns (such as ``de Sitter'' and ``Higgs boson'') come in such rigid combinations, such that further textual analysis can only benefit from representing them as compound mono-grams. 

Thus, we will further process the data by listing all the most common $2$-grams and then automatically hyphenating them into compound words, up to some cutoff in frequency. For example, as ``\verb|magnetic|'' and ``\verb|field|'' appear together frequently, we will replace this combination with ``\verb|magnetic-field|'', which is subsequently treated as a single unit. 
We note that even with all of the above, it is inevitable that some hyphenations or removals will be missed.
However, since we are doing largely a statistical analysis, such small deviations should not matter compared to the most commonly used words and concepts. The final output of this we will call {\bf cleaned data}. This process of iterative cleaning of the titles is itself illustrative; we leave further discussion to Appendix~\ref{ap:clean}.

As an example, our set of \hepth\ titles (cleaned) thus becomes a list of about $120,000$ entries, each being a list of words (both the mean and median is five words, down from the mean of 8 in the raw titles). 
A typical entry, in \mbox{\sf{Python}} format, would be (with our running example of~\cite{Witten:1991zd}),
\[
[ \text{`ground'}, \text{`ring'}, \text{`2dimensional'}, \text{`string-theory'} ]
\]
Note that the word ``of'' has been dropped because it is a trivial preposition, and the words ``two'' and ``dimensional'' have become joined to be ``2dimensional''. Both of these are done within the first steps of processing. Finally, the words ``string'' and ``theory'' have been recognized to be consistently appearing together by the computer in the final stages of cleaning the raw data, and the bi-gram has been replaced by a single hyphenated word.

We conclude this section by noting that steps \#4 and \#5 in the first stage of processing, and even the semi-automated merging of words that occurs in the cleaning of titles, requires a fair amount of field expertise to carry out successfully. This is not simply because the data contains \LaTeX\ markup language and an abundance of acronyms; it also requires a wide knowledge of the mathematical nomenclature of high energy physics, and the physical concepts contained therein. While it is possible, at least in principle, to imagine using machine learning algorithms to train a computer to recognize such compound $n$-grams as ``\verb|electric dipole moment|'', in practice this requires a fair amount of field expertise. To take another example, a computer will quickly learn that ``\verb|supersymmetry|'' is a noun, while ``\verb|supersymmetric|'' is an adjective. Yet the acronym ``\verb|susy|'' is used for both parts of speech in our community -- a bending of the rules that would complicate computational language processing.
{\it Thus it is crucial that this approach to computational textual analysis in high energy physics and mathematical physics be carried out by practitioners in the field, who also happen to have a rudimentary grasp of machine learning, computational linguistics and neural networks.}
The reader is also referred to an interesting recent work \cite{matrix} which uses matrix models to study linguistics as well as the classic works by \cite{paul1,paul2,paul3}.

\subsection{Frequency Analysis of \hepth}
\label{s:hepthfreq}

\begin{table}[t]
\centering
\begin{footnotesize}
\begin{tabular}{ | c | c | c || c | c | c | | c | c | c || c | c | c |} \hline
 \multicolumn{6}{|c||}{Raw Data}  &  \multicolumn{6}{|c|}{Cleaned Data}  \\
Rank & Word & Count & Rank & Word & Count &  Rank & Word & Count & Rank & Word & Count\\
\hline \hline
1 & of & 46,766 & 9 & quantum & 11,344 &  1 & model & 5,605 & 9 & field & 2,247  \\
2 & the & 43,713 & 10 & with & 10,003 &  2 & theory & 4,385 & 10 & equation & 2,245 \\
3 & and & 42,332 & 11 & field & 8,750 &  3 & black-hole & 4,231 & 11 & symmetry & 2,221 \\
4 & in & 39,515 & 12 & from & 8,690 &  4 & quantum & 4,007 & 12 & spacetime & 2,075 \\
5 & a & 17,805 & 13 & gravity & 7,347 &  5 & gravity & 3,548 & 13 & brane & 2,073 \\
6 & on & 16,382 & 14 & model & 6,942 &  6 & string & 3,392 & 14 & inflation & 2,031 \\
7 & theory & 13,066 & 15 & gauge & 6,694 &  7 & susy & 3,135 & 15 & gauge-theory & 2,014 \\
8 & for & 12,636 &  &  &   & 8 & solution & 2,596 &  &  &   \\
\hline
\end{tabular}
\caption{{\sf {\small The fifteen most common words in \hepth\ titles, in raw and clean data.
A graphical representation of this table, in terms of ``word-clouds'', is shown in Fig.~\ref{f:cloudhepth}.}}}
\label{tab:titlefreq}
\end{footnotesize}
\end{table}


Having raw and cleaned data at hand, we can begin our analysis with a simple frequency analysis of mono-grams and certain $n$-grams. For simplicity, we will here only discuss our primary focus, \hepth, leaving other sections of the \arXiv\ to Section~\ref{sec:compare}. 
The fifteen most common words in \hepth\ titles are given in Table~\ref{tab:titlefreq}. To understand the effect of our data cleaning process, we provide the counts for both the \textbf{raw} data, and the \textbf{cleaned} data. Note, for example, that the count for a word such as ``theory'' drops significantly after cleaning. In the clean data, the count on the word ``theory'' excludes all bi-grams involving this word that appear at least 50~times in \hepth\ titles, such as ``gauge-theory'', which appears 2014~times.

\begin{figure}[t]
\centerline{
(a)
\includegraphics[trim=0mm 0mm 0mm 0mm, clip, width=3in]{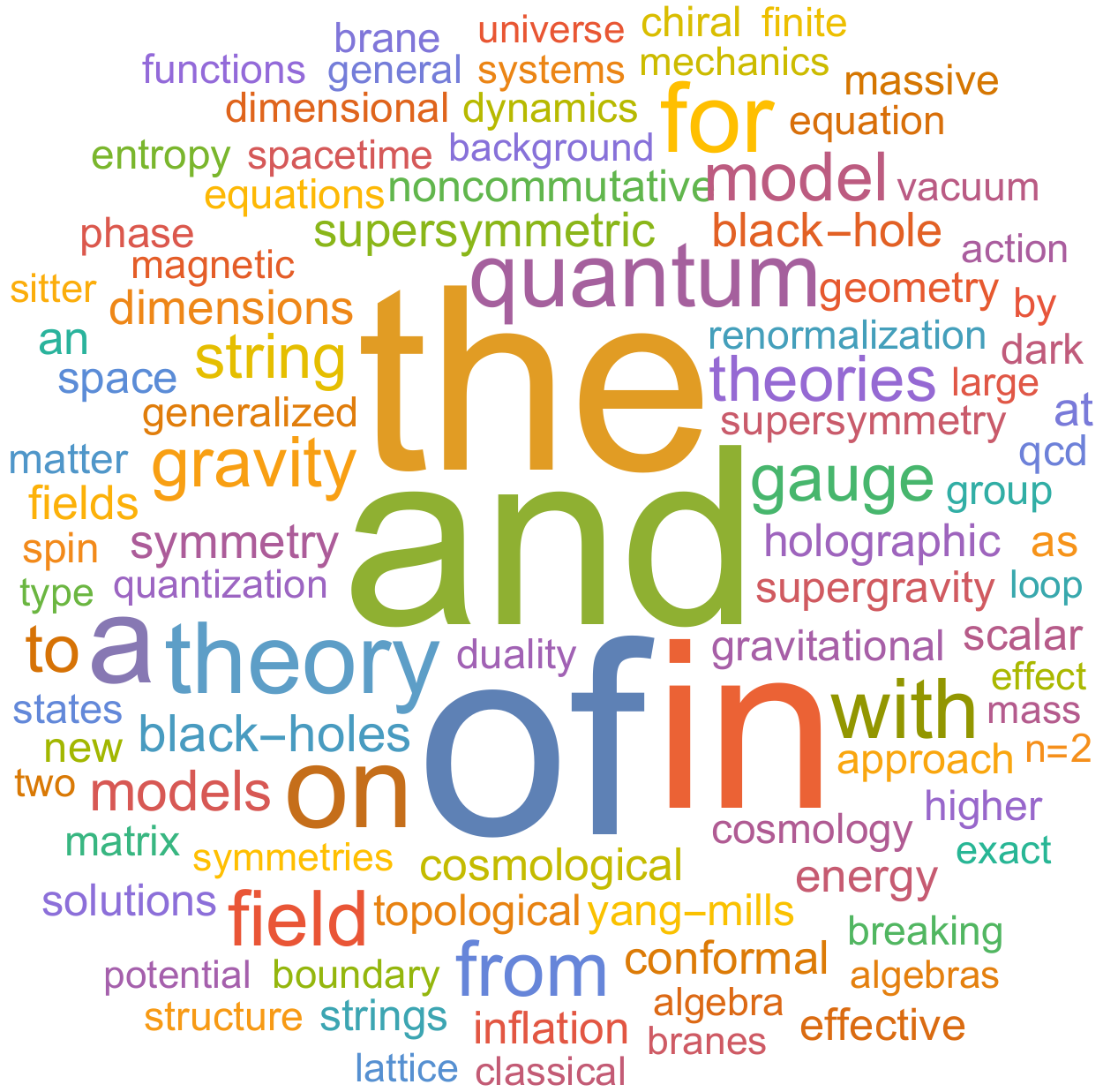}
(b)
\includegraphics[trim=0mm 0mm 0mm 0mm, clip, width=3in]{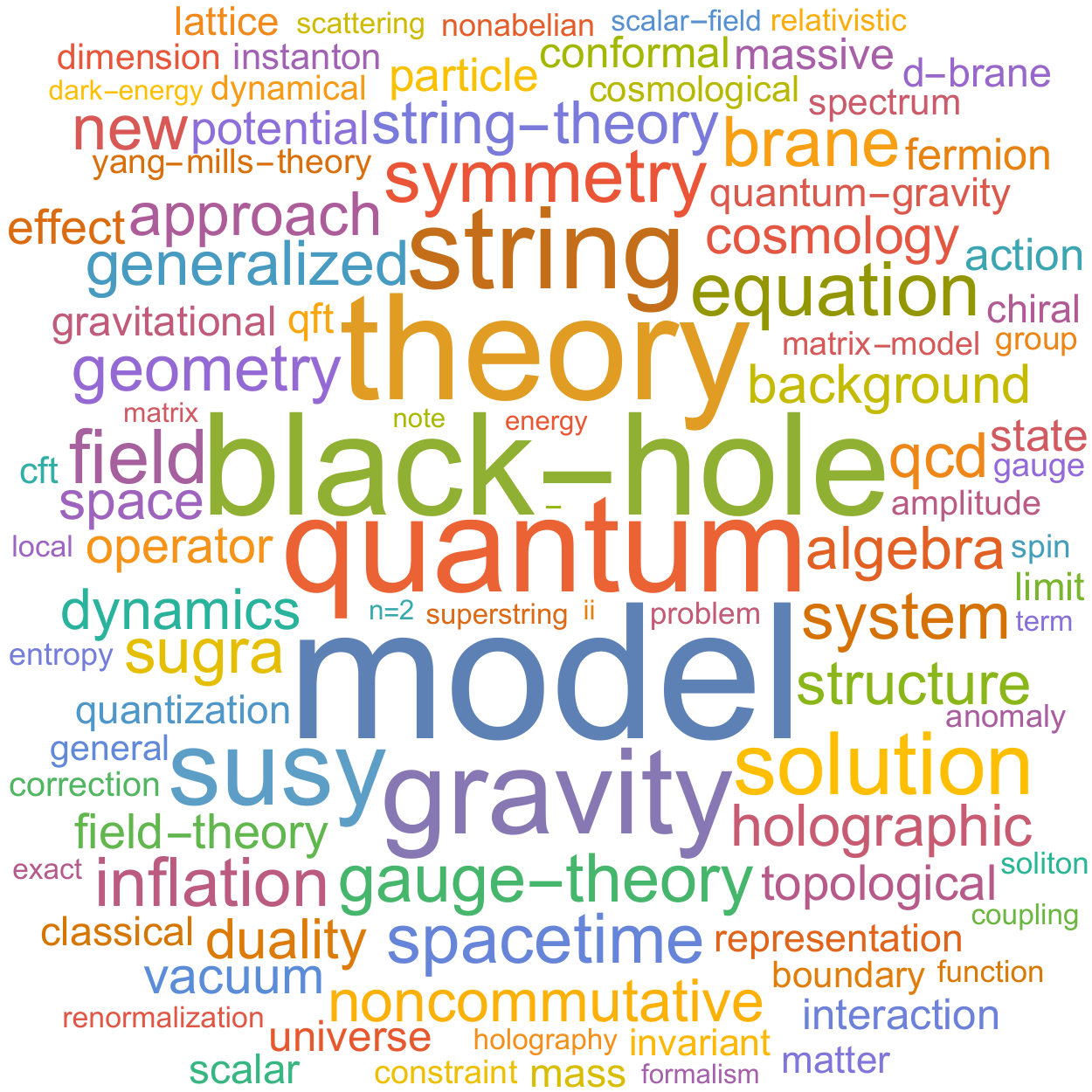}
}
\caption{{\sf {\small
The word clouds for (a) all raw titles (b) all cleaned titles within \hepth.
There is a total of 120,249 papers as of the end of 2017.
In the raw titles, there are 37,550 unique words and in the cleaned titles, 34,425.
\label{f:cloudhepth}}}}
\end{figure}

The standard method to present common words in natural language processing is called a {\bf word cloud}, where words are presented in aggregate, sized according to the frequency of their appearance.
We present the word clouds for the raw and cleaned titles in Figure~\ref{f:cloudhepth}.

In the above we encounter a first non-trivial observation. In the raw data, the word ``theory'' outnumbers the word ``model'' at nearly two-to-one. After the cleaning process, however, the order is reversed, and the word ``model''  emerges as the most common word. The explanation involves the grouping of individual words into bi-grams and tri-grams. In particular, the word ``theory'' ends up in common $n$-grams at a rate that is far larger than the word ``model'', which will turn out to be a major discriminatory observation that separates \hepth\ from other sections of the \arXiv.

Clearly, there is more discipline-specific contextual information in the cleaned data. For example, the most common technical word in all \hepth\ titles is ``black-hole''. The most common bi-gram involving the word ``theory'' is ``gauge-theory'', while ``string-theory'' appears with much lower frequency and is not one of the top-15 words, after cleaning. Note that not every instance of the word ``string'' appears in a common $n$-gram. Indeed, the word is more often used as an adjective in \hepth\ titles, as in ``string derived models'' or ``string inspired scenarios''.

For the abstracts of \hepth, we have only the raw data. Given the larger data set, and the prevalence of common, trivial, words, we here give the top~50 words in \hepth\ abstracts, together with their frequencies:
\begin{quote}
\{\text{the},1174435\}, \{\text{of},639511\}, \{\text{in},340841\}, \{\text{a},340279\}, \{\text{and},293982\}, \{\text{we},255299\}, \{\text{to},252490\}, \{\text{is},209541\}, \{\text{for},151641\}, \{\text{that},144766\}, \{\text{with},126022\}, \{\text{are},104298\}, \{\text{this},98678\}, \{\text{on},97956\}, \{\text{by},96032\}, \{\text{theory},86041\}, \{\text{as},78890\}, \{\text{which},71242\}, \{\text{an},68961\}, \{\text{be},66262\}, \{\text{field},65968\}, \{\text{model},50401\}, \{\text{from},49531\}, \{\text{at},46747\}, \{\text{it},46320\}, \{\text{can},46107\}, \{\text{quantum},44887\}, \{\text{gauge},44855\}, \{\text{these},39477\}, \{\text{also},36944\}, \{\text{show},35811\}, \{\text{theories},32035\}, \{\text{string},31314\}, \{\text{two},30651\}, \{\text{space},29222\}, \{\text{models},28639\}, \{\text{solutions},28022\}, \{\text{energy},27895\}, \{\text{one},27782\}, \{\text{study},26889\}, \{\text{gravity},25945\}, \{\text{fields},25941\}, \{\text{our},24760\}, \{\text{scalar},24184\}, \{\text{find},23957\}, \{\text{between},23895\}, \{\text{not},23273\}, \{\text{case},22913\}, \{\text{symmetry},22888\}, \{\text{results},22760\}
\end{quote}
%
The first technical words which appear are ``theory'', ``field'', ``model'', ``quantum'' and ``gauge''. As mentioned earlier, we have chosen not to ``clean'' the abstract data. It is interesting to note, however, that if the singular and plural form of words were combined, we would find ``theory'' and ``theories'' appearing 118,076 times, or almost once per abstract, with ``field''/``fields'' appearing in 91,909 abstracts, or just over 76\% of the total.


\begin{table}[t]
\centering
\begin{footnotesize}
\begin{tabular}{ | c || c | c || c | c |} \hline
& \multicolumn{2}{|c||}{Raw Data}  & \multicolumn{2}{|c|}{Cleaned Data}  \\
Rank & Bi-gram & Count & Bi-gram & Count \\
\hline \hline
1 & of the & 9287 & separation variable & 53 \\
2 & in the & 6418 & tree amplitude & 53 \\
3 & on the & 5342 & dark sector & 53 \\
4 & and the & 5118 & quantum chromodynamics & 53 \\
5 & field theory & 3592 & constrained system & 53 \\
6 & in a & 2452 & clifford algebra & 53 \\
7 & of a & 2111 & cosmological constraint & 53 \\
8 & for the & 2051 & black-hole information & 53 \\
9 & gauge theories & 1789 & black ring & 53  \\
10 & string theory & 1779 & accelerating universe & 53 \\
11 & field theories & 1468 & electroweak symmetry-breaking & 53 \\
12 & to the & 1431 & qcd string & 53 \\
13 & quantum gravity & 1412 & gravitational instanton & 52 \\
14 & gauge theory & 1350 & discrete torsion & 52 \\
15 & quantum field & 1242 & electric-magnetic duality & 52 \\
\hline
\end{tabular}
\caption{{\sf {\small The fifteen most common bi-grams in \hepth\ titles, in raw and clean data.}}}
\label{tab:titlebigram}
\end{footnotesize}
\end{table}

In addition to word frequency in the titles and abstracts, one could also study the $n$-grams. As it is clearly meaningless here to have $n$-grams cross different titles, we will therefore construct $n$-grams within each title, and then count and list all $n$-grams together. The fifteen most common bi-grams in \hepth\ titles are given in Table~\ref{tab:titlebigram}, again for raw and cleaned data.


There is little scientific content to be gleaned from the raw bi-grams, though we will find this data to be useful in Section~\ref{sec:compare}. 
In the cleaned bi-grams, many authors reference ``separation of variables'', ``tree amplitudes'', ``dark sectors'', ``quantum chromodynamics'', ``constrained systems'', ``Clifford algebras'', ``cosmological constraints'', ``black hole information'', ``black rings'', ``the accelerating universe'', etc.\ in their titles. It is clear to readers in the \hepth\ community that in the cleaned data set, many of the bi-grams would themselves be collective nouns if we imposed more rounds of automatic concatenation in the cleaning process.

For completeness, we conclude by giving the 50~most common bi-grams in the raw data for \hepth\ abstracts:
\begin{quote}
\{\{\text{of},\text{the}\},224990\}, \{\{\text{in},\text{the}\},115804\}, \{\{\text{to},\text{the}\},64481\}, \{\{\text{for},\text{the}\},49847\}, \{\{\text{on},\text{the}\},46444\}, \{\{\text{that},\text{the}\},44565\}, \{\{\text{of},\text{a}\},39334\}, \{\{\text{and},\text{the}\},38891\}, \{\{\text{can},\text{be}\},29672\}, \{\{\text{with},\text{the}\},29085\}, \{\{\text{show},\text{that}\},27964\}, \{\{\text{we},\text{show}\},24774\}, \{\{\text{in},\text{a}\},24298\}, \{\{\text{in},\text{this}\},23740\}, \{\{\text{it},\text{is}\},22326\}, \{\{\text{from},\text{the}\},20924\}, \{\{\text{by},\text{the}\},20634\}, \{\{\text{to},\text{a}\},18852\}, \{\{\text{as},\text{a}\},18620\}, \{\{\text{we},\text{find}\},17352\}, \{\{\text{we},\text{study}\},17113\}, \{\{\text{with},\text{a}\},16588\}, \{\{\text{field},\text{theory}\},16270\}, \{\{\text{we},\text{also}\},16184\}, \{\{\text{to},\text{be}\},15512\}, \{\{\text{is},\text{a}\},14125\}, \{\{\text{at},\text{the}\},13722\}, \{\{\text{terms},\text{of}\},13216\}, \{\{\text{for},\text{a}\},13112\}, \{\{\text{in},\text{terms}\},12573\}, \{\{\text{as},\text{the}\},12233\}, \{\{\text{study},\text{the}\},11638\}, \{\{\text{we},\text{consider}\},11481\}, \{\{\text{by},\text{a}\},11384\}, \{\{\text{of},\text{this}\},11372\}, \{\{\text{find},\text{that}\},11288\}, \{\{\text{on},\text{a}\},11272\}, \{\{\text{is},\text{the}\},11070\}, \{\{\text{in},\text{particular}\},10479\}, \{\{\text{which},\text{is}\},10478\}, \{\{\text{based},\text{on}\},10123\}, \{\{\text{we},\text{discuss}\},10051\}, \{\{\text{is},\text{shown}\},9965\}, \{\{\text{this},\text{paper}\},9871\}, \{\{\text{of},\text{these}\},9529\}, \{\{\text{between},\text{the}\},9480\}, \{\{\text{number},\text{of}\},9169\}, \{\{\text{string},\text{theory}\},8930\}, \{\{\text{the},\text{case}\},8889\}, \{\{\text{scalar},\text{field}\},8831\}
\end{quote}
Again, the most common bi-grams are trivial grammatical conjunctions. The first non-trivial combination is ``field theory'' and then, a while later, ``string theory'' and ``scalar field''. Indeed one would expect these to be the top concepts in abstracts in \hepth.
We remark that the current computer moderation of \arXiv\ uses full text, as it is richer and more accurate than titles and abstracts. and establishes a classifier which is continuously updated and uses adaptive length $n$-grams (typically up to n=4) \footnote{
We thank Paul Ginsparg for informing us of this.
}.

\section{Machine Learning hep-th}
\label{sec:deep}

\begin{table}[t]
\centering
\begin{footnotesize}
\begin{tabular}{ | c | c | c | c | c | c | c | c | c | c |} \hline
\multicolumn{10}{|c|}{Number of Unique Words Appearing at Least $N$ Times}  \\
$N=1$ & $N=2$ & $N=3$ & $N=4$ & $N=5$ & $N=6$ & $N=7$ & $N=8$ & $N=9$ & $N=10$ \\
\hline \hline
34425 & 16105 & 11642 & 9516 & 8275 & 7365 & 6696 & 6179 & 5747 & 5380\\
\hline
\end{tabular}
\caption{{\sf {\small Number of unique words appearing at least $N$ times in \hepth\ titles. The 16,105 words with at least two appearances were utilized as a training set for \wordvec for the purposes of the current section.}}}
\label{tab:unique}
\end{footnotesize}
\end{table}

Having suitably prepared a clean dataset, we then trained \wordvec\ on the collection of titles in \hepth. Given the typically small size of titles in academic papers, we chose a context window of length $5$. To minimize the tendency of the neural network to focus on outliers, such as words that very rarely appear, we dropped all words that appear less than twice in the data set for the purpose of training. As Table~\ref{tab:unique} indicates, that meant that a little over half of the unique words in the \hepth\ titles were not employed in the training.\footnote{Later, when we attack the classfication problem in Section~\ref{sec:class}, we will use all unique words in the training set.}
Finally, we follow standard practice in the literature by setting $N = 400$ neurons for the hidden layer, using the CBOW model. The result is that each of the non-trivial words is assigned a vector in $\IR^{400}$ so that the partition function~\eqref{Zcbow} is maximized.

\subsection{Word Similarity}
\label{sec:distance}

Once the word embedding has been established, we can form cosine distances in light of our definition of similarity and dissimilarity in Definition~\ref{sim}. 
This is our first glimpse into the ability of the neural network to truly capture the essence of syntax within the high energy theory community: to view which pairs of words the neural network has deemed ``similar'' across the entire corpus of \hepth\ titles.

Overall, we find that the neural network in \wordvec\ does an admirable job in a very challenging area of context. Consider the bi-gram ``\verb|super Yang-Mills|'', often followed by the word ``theory''. In step \#5 of the initial processing of the data, described in Section~\ref{s:clean}, we would have manually replaced this tri-gram with the acronym ``sym'', since ``SYM'' would be immediately recognized by most practitioners in our field as ``super Yang-Mills''. Thus the tri-gram ``\verb|supersymmetric Yang-Mills theory|'', a quantum field theory described by a non-Abelian gauge group, will be denoted as `\verb|sym|'.
Some representative word similarity measurements are
\begin{eqnarray}
d(\text{`sym'}, \text{`sym'}) &=& 1.0, \nonumber \\
d(\text{`sym'}, \text{`n=4'}) &=& 0.9763, \nonumber \\
d(\text{`sym'}, \text{`matrix-model'}) &=& 0.9569, \nonumber \\
d(\text{`sym'}, \text{`duality'}) &=& 0.9486, \nonumber \\ 
d(\text{`sym'}, \text{`black-hole'}) &=& 0.1567,\nonumber \\ 
d(\text{`sym'}, \text{`dark-energy'}) &=& -0.0188,\nonumber \\ 
d(\text{`sym'}, \text{`dark-matter'}) &=& -0.0370\, .
\label{dSYM}
\end{eqnarray}
The above means that, for example, ``sym'' is identical to itself (a useful consistency check), close to ``duality'', and not so close to ``dark-matter'', within our context windows. To practitioners in our field, these relative similarities would seem highly plausible. 

\begin{table}[t]
\centering
\begin{footnotesize}
\begin{tabular}{ | c || c | c || c | c |} \hline
Word & Most Similar & $d(w_1,w_2)$ & Least Similar & $d(w_1,w_2)$ \\
\hline \hline
model & theory & 0.7775 & entropy &  -0.0110 \\
theory & action & 0.7864 & holographic & 0.0079\\
black-hole & rotating & 0.9277 & lattice & 0.1332\\
quantum & entanglement & 0.8645 & sugra & 0.1880\\
gravity & massive-gravity & 0.8315 & $g^4$ & 0.0618 \\
string & string-theory & 0.9016 & approach & 0.0277 \\
susy & gauged & 0.9402 & energy & 0.0262 \\
solution & massive-gravity & 0.6900 & holographic & 0.0836\\
field & massless & 0.8715 & instanton & 0.0903\\
equation & bethe-ansatz & 0.8271 & matter & 0.0580\\
symmetry & transformations & 0.9286 & gravitational & 0.0095\\
spacetime & metric & 0.8560 & amplitude & 0.2502 \\
brane & warped & 0.9504 & method & 0.1706 \\
inflation & primordial & 0.9200 & cft & 0.1137 \\
gauge-theory & sym & 0.8993 & universe & 0.1507 \\
system & oscillator & 0.8729 & compactification & 0.1026 \\
geometry & manifold & 0.8862 & qcd & 0.1513 \\
sugra & gauged-sugra & 0.8941 & relativistic & 0.1939 \\
new & type & 0.8807 & state & -0.1240 \\
generalized & class & 0.9495 & effect & 0.0658\\
\hline
\end{tabular}
\caption{{\sf {\small The twenty most frequent words in \hepth\ titles. Included is the word with the largest and smallest values of the word-distance $d(w_1,w_2)$ (from equation~\ref{distance}), as computed by \wordvec.}}}
\label{tab:distance}
\end{footnotesize}
\end{table} 

We computed the similarity distance~(\ref{distance}) for all possible pairs of the 9516~words in \hepth\ titles which appear at least four times in the set. The twenty most frequent words are given in Table~\ref{tab:distance}, along with the word for which $d(w_1,w_2)$ is maximized, and where it is minimized. We refer to these words as the `most' and `least' similar words in the set.

Some care should be taken in interpreting these results. First, authors clearly use a different type of syntax when constructing a paper title than they would when writing an abstract, the latter being most likely to approximate normal human speaking styles. The semi-formalized rules that govern typical practice in crafting titles will actually be of interest to us in Section~\ref{sec:compare}, when we compare these rules across different sections of \arXiv.

The second caveat is that for two words to be considered similar, it will be necessary that the two words (1) appear sufficiently often to make our list, and (2) appear {\em together} in titles, within five words of one another, with high regularity. Thus we expect words like ``black hole'' and ``rotating'', or ``spacetime'' and ``metric'', to be naturally similar in this sense. How then should we interpret the antipodal word, which we designate as the ``least similar''? And what of the vast number of words whose cosine distance vanishes with respect to a particular word?

Recall the discussion in Section~\ref{s:word2vec}. The neural network establishes the vector representation of each word by attempting to optimize contextual relations. Thus two words will appear in the same region of the vector space if they tend to share many common words within their respective content windows. Inverting this notion, two words will be more likely to appear in antipodal regions of the vector space if the words they commonly appear with in titles are fully disjoint from one another. Thus ``sym'' (supersymmetric Yang-Mills theory) is not necessarily the `opposite' of ``dark matter'' or ``dark energy'' in any real sense, but rather the word ``sym'' tends to appear in titles surrounded by words like ``duality'' or ``matrix model'', which themselves rarely appear in titles involving the words ``dark matter'' or ``dark energy''. Thus the neural network located these vectors in antipodal regions of the vector space. 

The results of Table~\ref{tab:distance} also indicate that strictly negative values of $d(w_1,w_2)$ are, in fact, quite rare. So, for example, the word that is ``least similar'' to ``spacetime'' is ``amplitude'', with $d(\text{`spacetime'}, \text{`amplitude'}) = 0.25$. Part of the reason for this behavior is that Table~\ref{tab:distance} is presenting the twenty most frequent words in \hepth\ titles. Thus a word like ``spacetime'' appears in may titles, and develops a contextual affinity with a great many of the 9516~words in our dataset. As a consequence, no word in the \hepth\ title corpus is truly `far' from the word ``spacetime''.

\begin{figure}[t]
\centerline{
\includegraphics[trim=0mm 0mm 0mm 0mm, clip, width=3in]{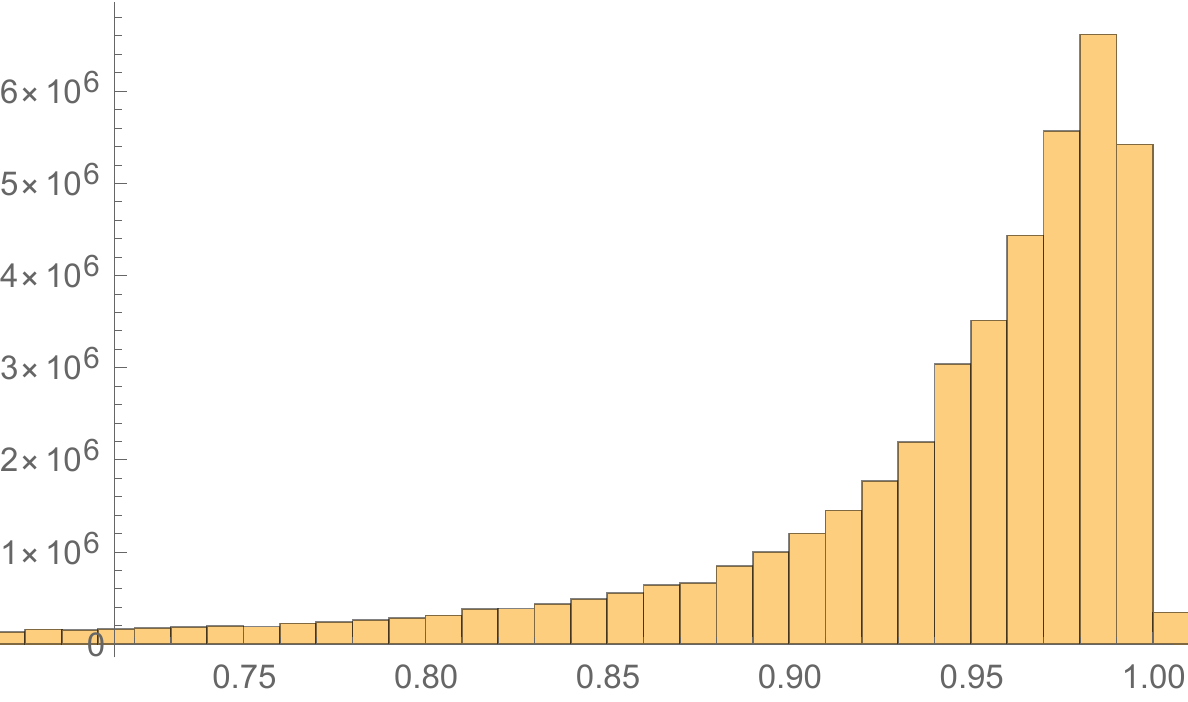}
\includegraphics[trim=0mm 0mm 0mm 0mm, clip, width=3in]{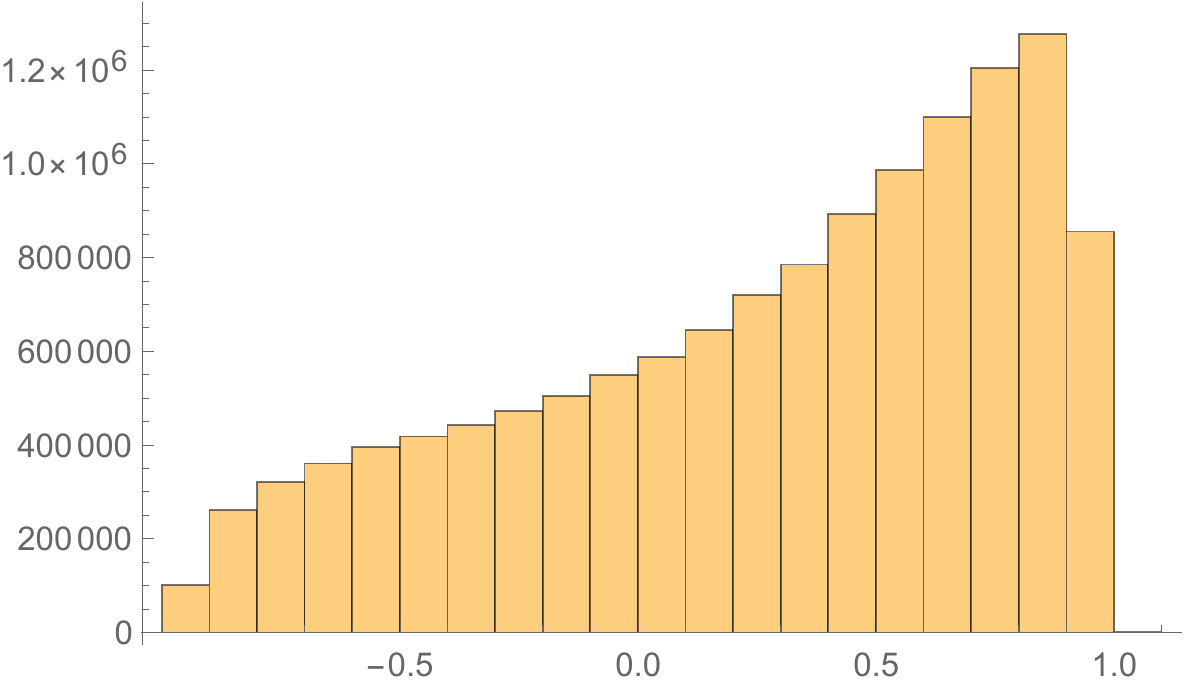}
}
\caption{{\sf {\small
Distribution of similarity distances, defined by~(\ref{distance}), for all pairs of words with at least four appearances in the corpus. The left panel gives the distribution for titles in \hepth. The right panel is the distribution for a similarly-sized collection of news headlines from {\em Times of India}. While both distributions show a clustering effect of words around certain dominant words, the shape of the distribution for \hepth\ shows a much tighter `conical' cluster than the hews headlines. In particular, we note the vertical scale in the two panels, in which negative similarity distances are essentially absent in \hepth\ titles, but reasonably frequent in the {\em Times of India} headlines. \label{f:pairs}}}}
\end{figure}

The relative lack of negative similarity distances, and the lop-sided nature of the vector space embedding produced by \wordvec, are striking for the \hepth\ titles. However, such behavior was noted in recent work by~Mimno and~Thompson on natural language processing with the skip-gram method~\cite{Mimno}. The authors observed that data sets tended to cluster in a cone around certain dominant words that appear frequently in context windows, such as the words like ``model'' and ``theory'' in our case. This is certainly the case with our data. A histogram of the $d(w_1,w_2)$ values for the 9516~words in \hepth\ titles which appear at least four times is given in the left panel of Figure~\ref{f:pairs}. The overwhelming majority of word pairs have similarity distances satisfying $d(w_1,w_2) \geq 0.9$, with a mean value given by
\begin{equation}  d(w_1,w_2)\big|_{\rm hepth} = 0.9257 \pm 0.0889 \, . \label{disthep}
\end{equation}

It might be thought that this extreme clustering is an artefact of the small size of the data set, or the restriction to words that appear at least four times. But {\em relaxing the cut-off to include all pairs of words that appear two or more times changes neither the average similarity distance nor the shape of the histogram}. As a control sample, we also trained \wordvec\ to produce a word embedding for the 22,950 unique words contained in 20,000 titles for news articles appearing in the {\it Times of India}~\cite{headline}. The same clustering affect occurs in this data sample, as can be seen by the histogram in the right panel of Figure~\ref{f:pairs}, but to a much more moderate extent. In fact, this data set contains a significant number of negative $d(w_1,w_2)$ values, with a mean value given by
\begin{equation}  d(w_1,w_2)\big|_{\rm headlines} = 0.2642 \pm 0.5171 \, . \label{disthead}
\end{equation}
We will return to this {\it Times of India} dataset in Section~\ref{sec:class}.

\subsection{Linear Syntactic Identities}
\label{s:hepth-linear}

With a measure of similarity, we can now seek examples of meaningful syntactic identities, analogous to~\eqref{king}.
This is done by finding the nearest vector to the vector sum/difference amongst word-vectors.
For example, we find that
\begin{equation}
\text{`holography'} + \text{`quantum'} + \text{`string'} + \text{`ads'} = \text{`extremal-black-hole'} \ .
\end{equation}
This is a correct expression, in that a hypothetical title containing the four words on the left hand side could plausibly contain the one on the right hand side.
This is very interesting because by {\it context} alone, we are uncovering the {\it syntax} of \hepth, where likely concepts appear together.
This is exactly the purpose of \wordvec, to attempt to study natural language through the proximity of context.

Another good example is
\begin{equation}
\text{`bosonic'} + \text{`string-theory'} = \text{`open-string'} \ .
\end{equation}
Of course, we need to take heed. It is {\em not} that the neural network has learned physics well enough to realize that the bosonic string has an open string sector; it is just that the neural network has learned to associate ``open-string'' as a {\it likely} contextual neighbor of ``bosonic'' and ``string theory''.
One could also add word-vectors to themselves (subtracting would simply give the word closest to the $0$-vector), such as (dropping the quotation marks for convenience):
\begin{verbatim}
gravity  +  gravity = massive-gravity
string  +  string = string-theory
quantum  +  quantum = quantum-mechanics
holography  +  holography = microscopic
\end{verbatim}
We will note that there is not much content to these identities.
It is not at all clear what scaling means in this vector space. 

We can systematically construct countless more ``linear syntactic identities'' from, say, our most common words.
For instance those of the form `a' + `b' = `c' include:
\begin{verbatim}
symmetry  +  black-hole  =  killing
spacetime  +  inflation  =  cosmological-constant
string-theory  +  spacetime  =  near-horizon
black-hole  +  holographic  =  thermodynamics
string-theory  +  noncommutative  =  open-string
duality  +  gravity  =  `d=5'
black-hole  +  qcd  =  plasma
symmetry  +  algebra  =  group
\end{verbatim}
The physical meaning of all of these statements are clear to a \hepth\ reader.

One can as well generate higher order examples  of the form `a' + `b' = `c` + `d', such as
\begin{verbatim}
field  +  symmetry  =  particle + duality
system  +  equation  =  classical + model
generalized  +  approach  =  canonical + model
equation  +  field  =  general + system
space  +  black-hole  =  geometry + gravity
duality  +  holographic  =  finite + string
string-theory + calabi-yau = m-theory + g2
\end{verbatim}
as well as even longer ones:
\begin{verbatim}
string-theory + calabi-yau + f-theory = orientifold
quiver + gauge-theory + calabi-yau = scft
brane + near-horizon - worldvolume = warped
\end{verbatim}
It is amusing that we can find many such suggestive identities.
On the other hand, we also find many statements which are simply nonsensical.

\section{Comparisons with Other \arXiv\ Sections}
\label{sec:compare}

In the previous section, we performed some basic descriptive analyses of the vocabulary of the \hepth\ community, demonstrated the power of \wordvec\ in performing textual analysis, illuminated the nature of the vector space created by the neural network output, and used this space to study word correlations and linear syntactic identities within the space generated by the corpus of all \hepth\ titles. We find these items interesting (or at least amusing) in their own right, and expect that there is much more that can be extracted by expert linguists or computer scientists.

As an application, in this section we would like to use the tools introduced above to perform an affinity analysis between the socio-linguistics of the \hepth\ community and sister communities that span theoretical high energy physics. 
There are many sections on \arXiv\ to which a great number of papers on \hepth\ are cross-listed. 
Similarly, authors who post primarily to \hepth\ often also post to other sections, and vice versa. 
This was particularly relevant prior to mid-2007, when each individual manuscript was referred to by its repository and unique number, and not solely by a unique number alone. As described in Section~\ref{s:sets}, the most pertinent physics sections besides \hepth\ are
\begin{itemize}
\item \hepph\ (high energy phenomenology);
\item \heplat\ (high energy lattice theory);
\item \mathph\ (mathematical physics);
\item \grqc\ (general relativity and quantum cosmology).
\end{itemize}
It is therefore potentially interesting to perform a cross-sectional comparative study.\footnote{
Note that of all the high energy sections, we have not included \mbox{\sf{hep-ex}}. This is because the language and symbols, especially the style of titles, of high-energy experimental physics are highly regimented and markedly different from the theoretical sections. This would render it an outlier in a comparative study of high-energy related \arXiv\ sections.}

What sorts of phenomena might we hope to identify by such a study? Loosely speaking, we might ask where authors in \hepth\ lie in the spectrum between pure mathematics and pure observation. Can we quantify such a notion by looking at the language used by authors in these various sections? The answer will turn out to be, in large part, affirmative. Intriguingly, we will see that the distinctions between the sub-fields is not merely one of {\em subject matter}, as there is a great deal of overlap here, but often it is encoded in the {\em manner} in which these subjects are described.

\subsection{Word Frequencies}
\label{s:monogram}
\begin{table}[t]
\centering
\begin{tabular}{| c | c || c | c || c | c |}
\hline \arXiv & No. of & Median & Mean  & Number of & Unique Word  \\
Section & Papers &  Length &  Length & Unique Words & Fraction \\
\hline \hline
\hepth & 120,249 & 5 & 5.08 & 34,425& 4.66\% \\
\hepph & 133,346 & 5 & 5.58 & 39,611 & 4.59\% \\
\heplat & 21,123 & 5 & 5.58 & 9,431 & 6.96\% \\
\grqc & 69,386 & 5 & 5.34 & 22,357 & 5.32\% \\
\mathph & 51,747 & 6 & 6.01 & 25,516 & 7.62\% \\
\hline
\end{tabular}
\caption{{\sf {\small Gross properties of the collection of titles of the five \arXiv\ sections studied in this paper, now after the cleaning process (described in Section~\ref{s:clean}) has been performed. The final column gives the number of unique words divided by the total number of words, after cleaning.}}}
\label{tab:alltitles}
\end{table} 

As in the previous sections, we first {\bf clean} the data to retain only relevant physics-concept related words. We again focus on the titles of papers in the five repositories. Some statistics for the data sets, in raw form, were given in Table~\ref{tab:wordcount}. After the cleaning process titles are typically shortened, and the number of unique words diminishes somewhat, as is shown in Table~\ref{tab:alltitles}. Again, the five repositories are roughly equal in their gross properties. We present the word-clouds for the four new repositories, after the cleaning procedure, in Table~\ref{f:cloud_all}, found in Appendix~\ref{ap:data}. 

At this point one could pursue an analysis for each repository along the lines of that in Section~\ref{sec:deep}. However, we will be less detailed in our study of the other four sections of \arXiv, as our focus in this section is two-fold: (1) to begin to understand similarities and differences between the authors in these five groupings, as revealed by their use of language, and (2) lay the groundwork for the classification problem that is the focus of Section~\ref{sec:class}.

\begin{table}[t]
\centering
\begin{footnotesize}
\begin{tabular}{| c | c || c | c || c | c || c | c || c | c |} \hline
\multicolumn{2}{|c||}{\hepth}& \multicolumn{2}{|c||}{\hepph}  & \multicolumn{2}{|c||}{\heplat} & \multicolumn{2}{|c||}{\grqc} & \multicolumn{2}{|c|}{\mathph}  \\
Word & \% & Word & \% & Word & \% & Word & \% & Word & \%  \\
\hline \hline
model & 0.80 & model & 0.84 & lattice & 1.91 & black-hole & 1.17  & model & 1.06 \\
theory & 0.62 & qcd & 0.58 & qcd & 1.43 & gravity & 0.96  & equation & 0.93 \\
black-hole & 0.60 & decay & 0.53 & lattice-qcd & 1.39 & spacetime & 0.75  & quantum & 0.92 \\
quantum & 0.57 & effect & 0.53 & model & 0.95 & model & 0.70  & system & 0.75 \\
gravity & 0.51 & lhc & 0.51 & quark & 0.58 & quantum & 0.58  & solution & 0.59 \\
string & 0.48 & dark-matter & 0.46 & theory & 0.54 & cosmology & 0.58  & theory & 0.49 \\
susy & 0.45 & neutrino & 0.41 & mass & 0.53 & universe & 0.55  & operator & 0.48 \\
solution & 0.37 & mass & 0.38 & fermion & 0.53 & theory & 0.52  & algebra & 0.45 \\
field & 0.32 & production & 0.38 & chiral & 0.51 & gravitational-wave & 0.50  & potential & 0.44 \\
equation & 0.32 & susy & 0.37 & meson & 0.41 & inflation & 0.46  & symmetry & 0.42 \\
\hline
\end{tabular}
\end{footnotesize}
\caption{{\sf {\small Top ten most frequent words in the titles of the five \arXiv\ sections, after cleaning. The percentage is the number of appearances of the particular word divided by the total of all words (including multiplicities).}}}
\label{tab:top10}
\end{table} 

We therefore begin with a focus on the most common `key' words in each section of \arXiv. In Table~\ref{tab:top10} we list the ten most common words in the titles of papers in each repository. We also give the overall word frequency for each word, normalized by the total number of words. Thus, for example, 1.91\% of all the words used in \heplat\ titles is the specific word ``lattice'', which is perhaps unsurprising. Indeed, the word frequencies in Table~\ref{tab:top10} are as one might expect if one is familiar with the field of theoretical high energy physics.

Many words appear often in several sections of \arXiv, others are common in only one section. Can this be the beginning of a classification procedure? To some extent, it can. For example, the word ``model'' is a very common word in all five sections, while ``theory'' fails to appear in the top ten words only for \hepph. However, this is deceptive, since it would be ranked number~11 if we were to extend the table. Despite such obvious cautions, there is still some comparative information which can be extracted from Table~\ref{tab:top10}. 

\begin{table}[t]
\centering
\begin{footnotesize}
\begin{tabular}{| c || c | c | c | c | c |} \hline
Top Word & \text{\hepth} & \text{\hepph} & \text{\heplat} & \text{\grqc} &
   \text{\mathph} \\ \hline
 \text{black-hole} & 3 & - & - & 1 & - \\
 \text{equation} & 10 & - & - & - & 2 \\
 \text{gravity} & 5 & - & - & 2 & - \\
 \text{mass} & - & 8 & 7 & - & - \\
 \text{model} & 1 & 1 & 4 & 4 & 1 \\
 \text{quantum} & 4 & - & - & 5 & 3 \\
 \text{solution} & 8 & - & - & - & 5 \\
 \text{susy} & 7 & 10 & - & - & - \\
 \text{theory} & 2 & - & 6 & 8 & 6 \\ \hline
\end{tabular}
\end{footnotesize}
\caption{{\sf {\small Words which are among the top ten most frequent for more than one \arXiv\ section, with the rank of the word in those sections.
The dashes `-' mean that the particular words has not made it into the top 10 of the specified \arXiv\ section.
}}}
\label{tab:top10combined}
\end{table} 

Consider Table~\ref{tab:top10combined}, in which we present only those words in Table~\ref{tab:top10} which appear in two or more sections of the \arXiv. We again see the universal importance of words like ``model'' and ``theory'', but we also begin to see the centrality of \hepth\ emerge. Indeed, it was for this reason that we chose to focus on this section of the \arXiv\ in the first place. Note that the words which are frequently found in \hepth\ tend to be frequent in other sections. Perhaps this represents some aspect of generality in the subject matter of \hepth, or perhaps it is related to the fact that amongst the other four sections, \hepth\ is far more likely to be the place where a paper is cross-listed than any of the remaining three sections.

What is more, even at this very coarse level, we already begin to see a separation between the more mathematical sections (\hepth, \grqc, and \mathph), and the more phenomenological sections (\hepph\ and \heplat). Such a divide is a very palpable fact of the sociology of our field, and it is something we will see illustrated in the data analysis throughout this section.

From the point of view of document classification, the simple frequency with which a word appears is a poor marker for the section of \arXiv\ in which it resides. In other words, if a paper contains the word ``gravity'' in its title, it may very likely be a \grqc\ paper, but the certainty with which a classifying agent -- be it a machine or a theoretical physicist -- would make this assertion would be low. As mentioned in Section~\ref{sec:tfidf}, term frequency-inverse document frequency (tf-idf) is a more nuanced quantity which captures the relative importance of a word of $n$-gram.

Recall that there are three important concepts when computing tf-idf values. First there are the words themselves, then there are the individual {\em documents} which, collectively, form the {\em corpus}. If our goal is to uncover distinctions between the five \arXiv\ sections using tf-idf values, it might make sense to take each paper as a document, with the total of all documents in a given section being the corpus. If we were studying the abstracts, or even the full text of the papers themselves, this would be the best approach. But as we are studying only the titles here, a problem immediately presents itself. 

After the cleaning process, in which small words are removed and common bi-grams are conjoined, a typical title is quite a small ``document''. As Table~\ref{tab:alltitles} indicates, the typical length of the document is five or six words. Thus, it is very unlikely that the {\em term frequency} across all titles will deviate greatly from the {\em document frequency} across the titles. This is borne out in the data. Taking the union of all words which rank in the top~100 in frequency across the five sections, we obtain 263~unique words. Of these 42.3\% never appear more than once in any title, in {\em any} of the five sections. For 63.5\% of the cases, the term frequency and document frequency deviate by no more than two instances, across {\em all five} \arXiv\ sections. Thus, tf-idf computed on a title-by-title basis is unlikely to provide much differentiation, as common words will have very similar tf-idf values.

\begin{table}[t]
\centering
\begin{footnotesize}
\begin{tabular}{| l || c | c | c | c | c |} \hline
Word & \text{\hepth} & \text{\hepph} & \text{\heplat} & \text{\grqc} &
   \text{\mathph} \\ \hline \hline
 \text{chiral-perturbation-theory} & 0.71 & 1.45 & 1.24 & 0. & 0.15 \\
 \text{cmb} & 1.36 & 1.43 & 0. & 1.4 & 0.49 \\
 \text{cosmological-model} & 4.73 & 0. & 0. & 5.86 & 0. \\
 \text{finite-volume} & 0.95 & 1.08 & 1.2 & 0. & 0.46 \\
 \text{ising-model} & 2.82 & 0. & 2.74 & 0. & 2.78 \\
 \text{landau-gauge} & 2.35 & 2.5 & 2.72 & 0. & 0. \\
 \text{lattice-gauge-theory} & 2.66 & 2.39 & 2.97 & 0. & 0. \\
 \text{lhc} & 1.19 & 1.87 & 0.69 & 0.9 & 0. \\
 \text{modified-gravity} & 2.89 & 2.42 & 0. & 3.22 & 0. \\
 \text{mssm} & 1.05 & 1.58 & 0.31 & 0.55 & 0. \\
 \text{neutrino-mass} & 2.24 & 3.54 & 0. & 0.35 & 0. \\
 \text{new-physics} & 0.92 & 1.57 & 0.31 & 0.15 & 0. \\
 \text{nucleon} & 0.78 & 1.65 & 1.35 & 0.15 & 0. \\
 \text{quantum-mechanics} & 1.49 & 1.15 & 0. & 1.28 & 1.37 \\
 \text{quark-mass} & 2.07 & 3.07 & 2.72 & 0. & 0. \\
 \text{scalar-field} & 1.51 & 1.34 & 0. & 1.56 & 1.12 \\
 \text{schrodinger-operator} & 0. & 0. & 0. & 0. & 10.08 \\
 \text{string-theory} & 1.67 & 1.28 & 0. & 1.28 & 0.97 \\
 \text{wilson-fermion} & 0. & 0. & 8.87 & 0. & 0. \\ \hline
\end{tabular}
\end{footnotesize}
\caption{{\sf {\small Term frequency-inverse document frequency (tf-idf) for certain key words. The corpus here is the set of all titles for all papers in all five sections of the \arXiv. These sections become the five documents of the corpus.}}}
\label{tab:tfidf}
\end{table} 

Therefore, we will instead consider an alternative approach. We let the corpus be all titles for all five sections of the \arXiv; in other words, we treat the entire (theoretical high energy) \arXiv\ as a single corpus. The collections of five titles form the five ``documents'' in this corpus. While this clearly implies some blurring of context, it gives a sufficiently large data set, document-by-document, to make tf-idf meaningful.

We provide the tf-idf values for a representative set of words in the five \arXiv\ sections in Table~\ref{tab:tfidf}. Recall that words that are extremely common in individual contexts (here, specifically, paper titles), across an entire \arXiv\ section, will have very low values of tf-idf. For example, words such as ``theory'' and ``model'' appears in all sections, and would therefore receive a vanishing value of tf-idf. Therefore, it is only illustrative to include words which do not trivially have a score of zero for all sections.
Our collection of titles is sufficiently large in all cases that there are no words which appear in {\em all} paper titles, even prior to the cleaning step, which removes small words like ``a'' and ``the''. Therefore, entries which are precisely zero in Table~\ref{tab:tfidf} are cases in which the word appears in {\em none} of the paper titles for that section of \arXiv.

So, for example, we see that, interestingly, the word ``schrodinger-operator'' appears only in \mathph\ but not in any of the others. This is, in fact, the word with the highest tf-idf score by far.  Of course, the Schr\"dinger equation is ubiquitous in all fields of physics, but only in \mathph\ -- and not even in \hepth\ -- is its operator nature being studied intensively. Likewise, the Wilson fermion is particular to \heplat. The word ``string-theory'' is mentioned in all titles except, understandably, in \heplat. 

Using machine learning techniques to classify \arXiv\ titles will be the focus of our next section, but we can take a moment to see how such an approach could potentially improve over a human classifier -- even one with expertise in the field. A full-blown tf-idf analysis would not be necessary for a theoretical physicist to surmise that a paper whose title included the bi-gram ``Wilson fermion'' is very likely from \heplat. But, surprisingly, having ``lattice gauge theory'' in the title is \textbf{not} a very good indicator of belonging to \heplat. Nor is it sufficient to assign \grqc\ to all papers with ``modified gravity'' in its title.

\subsection{Common Bi-grams}
\label{s:bigram}
As before, from mono-grams (individual words) we proceed to common $n$-grams. In this section we will concentrate on bi-grams for simplicity. Common 3-grams and 4-grams for the various sections can be found in Appendix~\ref{ap:data}. It is enlightening to do this analysis for both the cleaned data (for which we will extract subject content information), as well as for the raw data, which retains the conjunctions and other grammatically interesting words. The latter should give us an idea of the syntax of the language of high energy and mathematical physics across the disciplines.

\begin{table}[t]
\centering
\begin{footnotesize}
\begin{tabular}{| c | c | c | c | c |}
\hline
\hepth & \hepph & \heplat & \grqc & \mathph \\ \hline \hline
of the & of the & of the & of the & of the \\ \hline
in the & in the & lattice qcd & in the & on the \\ \hline
on the & dark matter & in the & on the & for the \\ \hline
and the & and the & on the & and the & in the \\ \hline
field theory & at the & the lattice & of a & and the \\ \hline
in a & on the & gauge theory & quantum gravity & of a \\ \hline
of a & in a & and the & dark energy & in a \\ \hline
for the & the lhc & in lattice & in a & to the \\ \hline
gauge theories & to the & from lattice & gravitational waves & for a \\ \hline
string theory & for the & lattice gauge & general relativity & on a \\ \hline
field theories & standard model & qcd with & for the & solutions of \\ \hline
to the & corrections to & at finite & scalar field & field theory \\ \hline
quantum gravity & production in & gauge theories & black-holes in & of quantum \\ \hline
gauge theory & production at & study of & gravitational wave & approach to \\ \hline
quantum field & from the & for the & with a & quantum mechanics \\ \hline
\end{tabular}
\end{footnotesize}
\caption{{\sf {\small The top~15 most commonly encountered bi-grams in the raw data, for each of the five sections of \arXiv. Bi-grams are listed from most frequent (top) to least frequent (bottom). \label{t:raw2grams}
}}}
\end{table}

\paragraph{Raw Data: }
The top~15 most commonly encountered bi-grams in the raw data, for each of the five sections of \arXiv, are presented in Table~\ref{t:raw2grams}, in descending order of frequency.
At first glance, the table may seem to contain very little distinguishing information. Clearly certain linguistic constructions, such as ``on the'' and ``on a'', are commonly found in the titles of academic works across many disciplines. One might be tempted to immediately remove such ``trivial'' bi-grams and proceed to more substantive bi-grams. But, in fact, there is more subtlety here than is immediately apparent. Let us consider the twelve unique, trivial bi-grams in the table above. They are given in Table~\ref{t:Trivial2grams} for each repository, with their ranking in the list of all bi-grams for that repository.

\begin{table}[t]
\centering
\begin{tabular}{| l || c | c | c | c | c |}
\hline
 & \multicolumn{5}{|c|}{\arXiv\ Repository Rank} \\ \hline \hline
Bi-gram & \hepth & \hepph & \heplat & \grqc & \mathph \\ \hline \hline
of the & 1 & 1 & 1 & 1 & 1 \\ \hline
in the & 2 & 2 & 3 & 2 & 4 \\ \hline
on the & 3 & 6 & 4 & 3 & 2 \\ \hline
and the & 4 & 4 & 7 & 4 & 5 \\ \hline
in a & 6 & 7 & 19 & 8 & 7 \\ \hline
of a & 7 & 16 & 35 & 5 & 6 \\ \hline
for the & 8 & 10 & 15 & 11 & 3 \\ \hline
to the & 12 & 9 & 21 & 16 & 8 \\ \hline
on a & 17 & 152 & 28 & 38 & 10 \\ \hline
from the & 19 & 15 & 27 & 27 & 74 \\ \hline
with a & 21 & 27 & 57 & 15 & 16 \\ \hline
at the & 87 & 5 & 97 & 114 & 224 \\ \hline
\end{tabular}
\caption{{\sf {\small Contextually `trivial' bi-grams across five \arXiv\ sections. The entry gives the rank of the bi-gram, in terms of the bi-gram frequency, after only removing capitalization. \label{t:Trivial2grams}
}}}
\end{table}

Three things immediately stand out. The first is the universal supremacy of the bi-gram ``of the''.  The second is the presence of ``at the'' in \hepph\ at a high frequency, yet largely absent from the other repositories. But this is clearly understood as the likelihood of \hepph\ titles to include phrases like ``at the Fermilab Tevatron'', or ``at the LHC'', in their titles. This is unique to \hepph\ among the five categories studied here. Finally, there is the construction ``on a'', which appears significantly only in \hepth\ and \mathph, and is very rare in \hepph\ titles. 

Again, this is easily understood, as the phrase ``on a'' generally precedes a noun upon which objects may reside. That is, plainly speaking, a surface. And the study of physics on surfaces of various sorts is among the most mathematical of the physical pursuits. So this meta-analysis of physics language syntax would seem to indicate a close affinity between \hepth\ and \mathph, and a clear distinction between \hepph\ and all the other theoretical categories. While this may confirm prejudices within our own fields, a closer inspection of these trivial bi-grams is warranted.

\begin{table}[t]
\centering
\begin{tabular}{| l || c | c | c | c | c |}
\hline
 & \hepth & \hepph & \heplat & \grqc & \mathph \\ \hline \hline
\hepth & 1 & 0.29 & 0.94 & 0.99 & 0.96 \\ \hline
\hepph & 0.29 & 1 & 0.39 & 0.38 & 0.12 \\ \hline
\heplat & 0.94 & 0.39 & 1 & 0.90 & 0.84 \\ \hline
\grqc & 0.99 & 0.38 & 0.90 & 1 & 0.95 \\ \hline
\mathph & 0.96 & 0.12 & 0.84 & 0.95 & 1 \\ \hline
\end{tabular}
\caption{{\sf {\small Treating the columns of Table~\ref{t:Trivial2grams} as vectors in  $\IR^{12}$, the cosine of the angles between the various vectors is given. \label{trivialcos}}}}
\end{table}

Considering Table~\ref{t:Trivial2grams} more seriously, we can treat the columns as vectors in a certain space of trivial bi-grams (not to be confused with word-vectors which we have been discussing). 
A measure of affinity between the authors of the various \arXiv\ sections would then be the cosine of the angle between these vectors. The value of these cosines is presented in Table~\ref{trivialcos}. 

What emerges from Table~\ref{trivialcos} is quite informative. It seems that the simplest of our community's verbal constructs reveal a great deal about how our colleagues organize into groups. Our central focus is the community of \hepth, and it is somewhat reassuring to see that the trivial bi-gram analysis reveals that this section has relatively strong affinity with all of the \arXiv\ sections studied. Yet there is clearly a break between \hepth\ and \hepph\, a distinction we will comment upon later. Clearly, some of this is driven by the ``at the'' bi-gram, suggesting (quite rightly) that \hepph\ is the most experimentally-minded of the \arXiv\ sections studied here. But even if this particular bi-gram is excluded from the analysis, \hepph\ would still have the smallest cosine measure with \hepth.

Clearly, the trivial bi-gram analysis suggests that our group of five sections fragments into one section (\hepph), and the other four, which cluster rather tightly together. Among these remaining four, \heplat\ is slightly the outlier, being more closely aligned with \hepth\ than the other sections. Most of these relations would probably not come as a surprise to authors in the field, but the fact that the computer can make distinctions in such a specialized sub-field, in which even current practitioners would have a difficult time making such subtle differentiation, is intriguing.

\paragraph{Cleaned Data: }
After the cleaning process, which includes the two rounds of computer generated word concatenations, described in Appendix~\ref{ap:clean}, the majority of the most common bi-grams in each section of the \arXiv\ will have been formed into single words. What remains reveals something of the specific content areas unique to each branch of theoretical particle physics. The 15~most frequent bi-grams after cleaning are given in Table~\ref{t:clean2gram}.

\begin{table}[t]
\begin{footnotesize}
\centering
\begin{tabular}{| l | l |}
\hline
\text{\hepth}&
\{\text{separation},\text{variable}\}, \{\text{tree},\text{amplitude}\}, \{\text{dark},\text{sector}\}, \{\text{quantum},\text{chromodynamics}\}, \\
& \{\text{constrained},\text{system}\}, \{\text{clifford},\text{algebra}\}, \{\text{cosmological},\text{constraint}\}, \{\text{black-hole},\text{information}\}, \\
& \{\text{black},\text{ring}\}, \{\text{accelerating},\text{universe}\}, \{\text{electroweak},\text{symmetry-breaking}\}, \{\text{qcd},\text{string}\}, \\
& \{\text{gravitational},\text{instanton}\}, \{\text{discrete},\text{torsion}\}, \{\text{electric-magnetic},\text{duality}\}
\\ \hline
\text{\hepph}&
\{\text{first-order},\text{phase-transition}\}, \{\text{chiral-magnetic},\text{effect}\}, \{\text{double-parton},\text{scattering}\}, \\
& \{\text{littlest-higgs-model},\text{t-parity}\}, \{\text{momentum},\text{transfer}\}, \{\text{extensions},\text{sm}\}, \{\text{magnetic},\text{catalysis}\}, \\
& \{\text{jet},\text{substructure}\}, \{\text{matter},\text{effect}\}, \{\text{energy},\text{spectrum}\}, \{\text{spin-structure},\text{function}\}, \\
& \{\text{equivalence},\text{principle}\}, \{\text{light-scalar},\text{meson}\}, \{\text{au+au},\text{collision}\}, \{\text{searches},\text{lhc}\}
\\ \hline
\text{\heplat}&
\{\text{flux},\text{tube}\}, \{\text{perturbative},\text{renormalization}\}, \{\text{imaginary},\text{chemical-potential}\}, \{\text{ground},\text{state}\}, \\
& \{\text{gluon},\text{ghost}\}, \{\text{electroweak},\text{phase-transition}\}, \{\text{string},\text{breaking}\}, \{\text{physical},\text{point}\}, \\
& \{\text{2+1-flavor},\text{lattice-qcd}\}, \{\text{lattice},\text{action}\}, \{\text{2+1-flavor},\text{qcd}\}, \{\text{random-matrix},\text{theory}\}, \\
& \{\text{effective},\text{action}\}, \{\text{screening},\text{mass}\}, \{\text{chiral},\text{transition}\}
\\ \hline
\text{\grqc}&
\{\text{fine-structure},\text{constant}\}, \{\text{extreme-mass-ratio},\text{inspirals}\}, \{\text{closed-timelike},\text{curves}\}, \{\text{bulk},\text{viscosity}\}, \\
 & \{\text{born-infeld},\text{gravity}\}, \{\text{dirac},\text{particle}\}, \{\text{ds},\text{universe}\}, \{\text{einstein-field},\text{equation}\}, \{\text{fundamental},\text{constant}\}, \\
 & \{\text{topologically-massive},\text{gravity}\}, \{\text{bose-einstein},\text{condensate}\}, \{\text{higher-dimensional},\text{black-hole}\}, \\
 & \{\text{hamiltonian},\text{formulation}\}, \{\text{static},\text{black-hole}\}, \{\text{generalized-second},\text{law}\}
\\ \hline
\text{\mathph}&
\{\text{time},\text{dependent}\}, \{\text{external},\text{field}\}, \{\text{thermodynamic},\text{limit}\}, \{\text{long},\text{range}\}, \{\text{variational},\text{principle}\}, \\
 & \{\text{loop},\text{model}\}, \{\text{minkowski},\text{space}\}, \{\text{fokker-planck},\text{equation}\}, \{\text{characteristic},\text{polynomials}\}, \\
 & \{\text{hamiltonian},\text{dynamics}\}, \{\text{integral},\text{equation}\}, \{\text{configuration},\text{space}\}, \{\text{lattice},\text{model}\}, \\
 & \{\text{constant},\text{curvature}\}, \{\text{gaussian},\text{free-field}\}
\\ \hline
\end{tabular}
\end{footnotesize}
\caption{{\sf  {\small Most common bi-grams in cleaned data, after two rounds of automated concatenation. \label{t:clean2gram}}}}
\end{table}

While these bi-grams certainly capture important areas of theoretical physics research in each section of the \arXiv\, they are somewhat deceiving. For example, the ``chiral magnetic effect'' -- a phenomenon of induced chiral behavior in a quark-gluon plasma -- has been the subject of study of roughly five papers per year in \hepph\ over the last decade. But it would be wrong to suppose that it is more commonly studied than ``searches (at the) lhc'', or ``extensions (to the) sm'', where the acronyms stand for Large Hadron Collider and Standard Model, respectively.

In this case, we see a potential drawback of the automated concatenation described in Appendix~\ref{ap:clean}: whereas tightly related topics will naturally be grouped together, some of the diversity of subject matter in each section of the \arXiv\ will be obscured. So, for example, in step \#5 in Section~\ref{s:clean} we turn common compound expressions like ``operator product expansion'' into the acronym ``ope''. In addition, a handful of very common bi-grams were hyphenated, such as ``dark matter'' becoming ``dark-matter''. Cosmological dark matter is a topic of investigation in \hepph\ which appears to be totally missing in Table~\ref{t:clean2gram}! However, what has happened is that various sub-categories of postulated dark matter candidates have been concatenated in the automated steps which follow, producing ``cold-dark-matter'', ``warm-dark-matter'', ``fuzzy-dark-matter'', etc. The total frequency of appearance for ``dark matter'' itself is thus distributed across many `words' in the total corpus.

To take another example, particle physicists of a certain age will remember the flurry of papers appearing (primarily) in \hepph\ in the early 2000's with the bi-gram ``little Higgs'' in the title. Indeed, for a certain period of time it seemed that {\em every} paper in \hepph\ concerned this alternative to the traditional Higgs mechanism of the Standard Model. From this were spun many off-shoots with their ever-more creative names: ``littler Higgs'', ``littlest Higgs'' (a hint of which appears in Table~\ref{t:clean2gram}), ``fat Higgs'', ``thin Higgs'', etc. Where are all the ``little Higgs'' papers in our study?

\begin{table}[t]
\centering
\begin{tabular}{| l || c | c | c | c | c |}
\hline
Key Word & \hepth & \hepph & \heplat & \grqc & \mathph 
\\ \hline \hline
theory & 290 & 72 & 144 & 141 & 131 \\ 
model & 259 & 251 & 185 & 178 & 228 \\ \hline
Higgs & 21 & 123 & 24 & 11 & 5 \\
dark & 3 & 8 & -- & 8 & 1 \\
black & 6 & 1 & -- & 9 & 3 \\
natural & 2 & 2 & -- & 1 & -- \\
conformal & 33 & 5 & 14 & 27 & 25 \\ \hline
\end{tabular}
\caption{{\sf {\small Number of unique bi-grams formed with seven key words, across the five \arXiv\ sections. Frequencies are computed from \textbf{processed} data (small words are removed but automated concatenation is not performed). \label{t:common2grams}}}}
\end{table}

The answer, as Table~\ref{t:common2grams} shows, is that they are still there -- only hidden. In the table, we work with \textbf{processed} data, which (as described in Section~\ref{s:clean}) is data in which the first five steps of the cleaning process -- through the removal of small words -- is performed, but {\em prior} to the automated concatenation of common bi-grams. What is given in Table~\ref{t:common2grams} is the number of unique bi-grams constructed with seven very common key words, for each of the five sections. To construct the Table above, we only considered the top-6000 bi-grams in each section, ranked by frequency of appearance.

First we remark on the sheer number of bi-grams formed with ``theory'' and ``model''. Prior to automated concatenation, ``model'' far outstrips ``theory'' in \hepph, and to a lesser extent in \mathph. After the automated concatenation ({\em i.e.} in the cleaned data), ``model'' will be one the most common, or nearly most common, words in all five sections, as shown in Table~\ref{tab:top10}. As we remarked in Section~\ref{s:hepthfreq}, it is this automated concatenation that will eventually reduce the frequency of the stand-alone word ``theory'' in \hepth.

The other five words reveal a great deal about the subject matter of the five \arXiv\ sections, as well as demonstrating the creativity of the high energy theory community in creating new bi-grams. This is particularly so, as expected, in \hepph's treatment of the word ``Higgs'', in which an astounding 123~bi-grams are identified involving this word, including everything from ``abelian Higgs'' to ``Higgs vacuum''. The distinction between the phenomenological (\hepph\ and \heplat) and more formal sections is evident in the number of bi-grams involving ``conformal'', with \hepth\ leading the group. It seems \hepph\ concerns itself more with ``dark'' objects (``atoms", ``forces", ``gauge", ``photon", ``radiation", ``sector", ``side") than ``black'' ones; it is vice versa for \hepth\ (``black'' plus ``brane", ``hole'', ``objects", ``p-brane", ``ring", ``string"); and \grqc\ is concerned with both in equal measure. Finally, despite the frequent use of ``natural'' in our community -- and its ubiquitous adjectival noun form, ``naturalness'' -- this word appears paired in a bi-gram with only two words with any great regularity: ``inflation'' and ``susy (supersymmetry)''.

\subsection{Comparative Syntactic Identities}

We conclude this amusing bit of navel-gazing with a comparison of the five \arXiv\ sections using the vector space word embeddings produced by \wordvec. As in Section~\ref{sec:deep}, we train \wordvec\ on the five lists of words formed from the titles of the five sections, after the full cleaning procedure is performed.  What we are seeking is differences in the way certain common words are represented in the five constructed vector spaces. 

It is intriguing to consider how linear syntactic identities, such as those identified in Section~\ref{s:hepth-linear}, are modified when we map words from one vector space embedding to another. It would be particularly interesting to see if, in different sections of the \arXiv, the same left-hand side of a linear syntactic identity `a' + `b' = `c' leads to completely different `c' in different spaces. This would be indicative of the very nature of the terminologies of the fields.

Many possible `a' + `b' pairs can be constructed, though very frequently one of the members of the (`a',`b') pair happens to be seldom used in at least one of the five \arXiv\ sections. One word which appears frequently with many partners is the word `spin'. Some syntactic identities using this word include:
\begin{verbatim}
hep-th:  spin  +  system  =  free
hep-ph:  spin  +  system  =  1/2
hep-lat:  spin  +  system  =  ising
gr-qc:  spin  +  system  =  initial-data
math-ph:  spin  +  system  =  charged

hep-th:  spin  +  dynamics  =  point
hep-ph:  spin  +  dynamics  =  chromofield
hep-lat:  spin  +  dynamics  =  geometry
gr-qc:  spin  +  dynamics  =  orbiting
math-ph:  spin  +  dynamics  =  interacting

hep-th:  spin  +  effect  =  electron
hep-ph:  spin  +  effect  =  role
hep-lat:  spin  +  effect  =  bound-state
gr-qc:  spin  +  effect  =  detector
math-ph:  spin  +  effect  =  glass
\end{verbatim}
What these identities reveal is that `spin' is used in very general contexts in both \hepth\ and \hepph, but in rather more specific contexts in the other three sections. So, for example, in \grqc\ a topic of inquiry may be the dynamics of spinning objects orbiting a black hole, whereas the thermodynamics and stability properties of spin glasses has been a common topic of research in \mathph.

We conclude this section with the syntactic identity formed from perhaps the two most fundamental concepts in theoretical particle physics: ``field theory'' (the tool with which all of our calculations are performed), and ``scattering'' (the primary observable we labor to compute). Across the five disciplines, these two concepts generate the following:
\begin{verbatim}
hep-th:  scattering  +  field-theory  =  green-function
hep-ph:  scattering  +  field-theory  =  dispersion-relation
hep-lat:  scattering  +  field-theory  =  wave-function
gr-qc:  scattering  +  field-theory  =  bms
math-ph:  scattering  +  field-theory  =  internal
\end{verbatim}
Note that ``bms'' here refers to the Bondi-Metzner-Sachs (BMS) group, a symmetry property of asymptotically-flat Lorentzian spacetimes, which was shown by Strominger to be a symmetry of the $S$-matrix in perturbative quantum gravity~\cite{Strominger:2013jfa}.

\section{Title Classification}
\label{sec:class}

The previous sections would be of some mild interest to theoretical particle physicists who routinely publish in the areas covered by our five \arXiv\ sections. But for practicing data scientists, what is of interest is the ability of the word embeddings to generate a classification algorithm that accurately and efficiently assigns the proper \arXiv\ section to a given paper title. In our context, the question is naturally
\begin{quote}
{\it
Question:  Given a random title (not even necessarily a legitimate physics title), can a neural network decide to which section on the \arXiv\ is it likely to belong?
}
\end{quote}
It turns out that one of the most powerful uses for \wordvec\ is classification of texts~\cite{textclass}. Indeed, document classification is one of the key tasks a neural network is asked to perform in natural language processing generally.

Classification is the canonical problem in supervised machine learning, and there are many possible approaches. The preceding sections have provided us with insights that will prove of value. In particular, `cleaning' the data allowed us to examine certain contextual relations and make more meaningful frequency statements about certain concepts. Nevertheless, it eliminated some information that may be useful, such as the pattern of small words like conjunctions, and it tended to bury certain key descriptors (like `dark') by embedding them in multiple hyphenated words. We will therefore train our classifier with \textbf{raw} data, with the only processing being the removal of upper-case letters.

The algorithm we will choose is the {\it mean word-vector} method~\cite{textclass}, and it proceeds as follows:
\begin{itemize}
\item Combine all titles from all the relevant sections from \arXiv; establish a word-vector for each word using \wordvec's CBOW neural network model. This gives a single vector space $V$ consisting of many 400-vectors (recall that our convention is such that each word-vector is an element of $\IR^{400}$); 
\item Subsequently, for each title, considered as a list of words, take the component-wise mean of the list of word-vectors; thus each title is associated with a single vector in  $V$;
\item Establish {\it labelled data} $D$ consisting of entries (Title$_1$, $i$), (Title$_2$, $i$), \ldots, (Title$_k$, $j$), (Title$_{k+1}$, $j$), \ldots
over all \arXiv\ sections under consideration; Here $i,j,\ldots \in \{1,2,3,4,5\}$ specifies the 5 sections, respectively, \hepth, \hepph, \heplat, \grqc~and \mathph;
\item Construct a ``training set'', say 7000 random samples from $D$ and use one's favourite classifier to train this sample;
\item Construct a ``validation set'', a complementary set of, say 13,000 samples from $D$, to ensure that the classifier has {\it not seen} these before. 
\item Predict to which \arXiv\ section each title in the validation set should belong, and check the veracity of that prediction.
\end{itemize}


A few remarks are in order. First, we emphasize: for the input titles, we did {\em not} perform any cleaning of the data, and the only processing that was done was to put all letters to lower-case. This is important because we wish to keep all grammatical and syntactical information, including conjunctions and (in)definite articles, etc., which could be indicative of the stylist choice in different sections of the \arXiv.
Moreover, that during the training of \wordvec\ we keep all words (instead of a cut-off at at least frequency four, as was done in Section \ref{sec:deep} for the cleaned data). This is because we will establish as large a vocabulary as possible in order to (1) ensure the labelling of each title; and (2) accommodate new unencountered titles.

Now, in the labelling of the titles, it may at first appear that by averaging over the word-vectors in a title, one loses precious information such as the ordering of words.
However, from similar studies in the literature~\cite{textclass}, such a seemingly crude method (or variants such as taking component-wise max/min) is actually extremely effective.
Furthermore, an often-used method is to {\it weight} the words in different documents (here, \arXiv\ sections) with tf-idf, but we will not do so here since, as was described in Section~\ref{s:monogram}, the vast majority of tf-idf values will be vanishing for the five ``documents'' crafted from the bag of all titles.

Finally, of the choices of classifiers, we use a support vector machine (SVM). Of course, other classifiers and neural networks can also be used, but SVMs are known to be efficient in discrete categorization (cf.~\cite{book}), thus we will adhere to this for this paper.

\subsection{Prediction Results}

Following the above algorithm, and applying it to our five \arXiv\ sections, we can establish the so-called {\bf confusion matrix} $\cM$, the $(i,j)$-th entry of which is established in the following way.
Comparing to a title coming from section $i$, suppose the SVM predicted section $j$, then we add 1 to $\cM_{ij}$.
By construction, $\cM$ is not necessarily symmetric.
Furthermore, each row sum is equal to the validation size, since the SVM is trained to slot the result into one of the categories;
column sums, on the other hand, need not have any constraints -- for instance, everything could be \mbox{(mis-)classified} into a single category.
In an ideal situation of perfect prediction by the SVM, $\cM$ would, of course, be the identity matrix, meaning there are no mismatches at all.

Using the ordering of the sections as $(1,2,3,4,5) =$ (\hepth, \hepph, \heplat, \grqc, \mathph), we find the confusion matrix to be:
\begin{equation}\label{CM}
\begin{array}{c|ccccc}
$\backslashbox{Actual}{\wordvec + SVM}$
    & 1 & 2 & 3 & 4 & 5 \\ \hline
 1 & 5223 & 844 & 1132 & 3122 & 2679 \\
 2 & 1016 & 8554 & 1679 & 1179	 & 572 \\
 3 & 977 & 1466	 & 9411 & 188 & 958 \\
 4 & 1610 & 566 & 128 & 9374 & 1322 \\
 5 & 1423 & 279 & 521 & 1010 & 9767 \\
\end{array} 
\qquad \qquad
\left\{
\begin{array}{rcl}
1&:&\hepth \\
2&:&\hepph \\
3&:& \heplat\\
4&:& \grqc \\
5&:& \mathph\, . \\
\end{array}
\right.
\end{equation}
%
%
%
%
We see that the classification is actually quite good, with confusion largely diagonal, especially in the last four sections, achieving around 70\% accuracy. The overall {\bf accuracy} is defined as the sum of the true positives (diagonal entries) divided by the total number of entries in the matrix (65,000), which yields 65.1\%. Let us take a moment to put this accuracy rate into perspective. In the years since the arrival of \wordvec, in 2013, a number of papers have appeared that aim to improve upon the original technique. In this literature, the goal is generally to classify documents into categories that are well-defined and highly distinct. For example, one might ask the classifier to distinguish whether a Wikipedia article~\cite{wiki} is about an office-holder or an athlete, or whether a thread on Yahoo!~Answers~\cite{yahoo} is about ``science \& math'' or ``business \& finance'' (c.f. the discussion in Zhang et al.~\cite{Zhang}). 
Such trials have become somewhat standardized into benchmark tests in the NLP community. For example, Zhang et al. demonstrated that \wordvec\ was able to accomplish the above-mentioned tasks of sorting DBPedia articles and Yahoo! Answers threads, with an accuracy of 89\% and 56\%, respectively.
In those cases, the classifier is generally given a much larger sample of words to address, and the task at hand is such that one expects a human classifier to perform the task with very high fidelity. In our test, however, it would be difficult for active researchers in high energy theory -- frequent contributors to the \arXiv\ -- to achieve even 65\% accuracy in sorting papers solely by title alone. One expects even better classification results can be achieved if one were to consider full abstracts, or even the entire text of papers \footnote{
Physicists distinguish titles of genuine papers posted on \hepth\ from fake titles generated using a context free grammar~\cite{snarxiv1} successfully only $59$\% of the time~\cite{snarxiv2}
}.

The mis-classifications are, themselves, very indicative of the nature of the sub-fields.
For example, \hepth\ (1st entry) is more confused with \grqc\ (4th entry) and \mathph\ (5th entry) than any other mis-classifications.
Indeed, this reflects that high-energy theory is closer to these two fields than any other field is close to another.
The asymmetry is also relevant: \hepth\ is more frequently confused (24.0\%) with \grqc\  than vice versa (12.4\%). This is because the string theory community that populates \hepth\ derives from both a particle physics and a gravity tradition, whereas the \grqc\ community has different historical roots.
This is similarly true for \hepth\ and \mathph. 
The cultural origins of the communities can already be deduced from this analysis.

Similarly, the two more phenomenological sections, \hepph\ and \heplat\, form a connected and isolated $2\times 2$ block, more often confused with one-another than any of the other three sections. And again, of the remaining three sections, \hepth\ stands out for mis-classification. As mentioned earlier, this suggests the centrality of \hepth\ in the organization of this community of particle physicists. One can make this analysis more precise by reducing the $5\times 5$ confusion matrix in~(\ref{CM}) to a binary $2\times 2$ form, in which we group \hepth, \grqc, and \mathph\ as ``formal'' sections, and group \hepph\ and \heplat\ as ``phenomenological'' sections. In this form, the binary classification matrix is
\begin{equation}  \cM = \left(
\begin{array}{cc}
35530 & 3470 \\
4890 & 21120 \\
\end{array}
\right)\, , 
\label{thvsph}
\end{equation}
which corresponds to an accuracy, in executing this binary classification, of 87.1\%, which is a remarkable success rate for such a subtle classification task.

\begin{table}[t]
\begin{footnotesize}
\centering
\begin{tabular}{| c | c | l |} \hline
True& Predicted& Title \\
\hline \hline
\hepth & \grqc & `string dynamics in cosmological and black-hole backgrounds: the null string expansion' \\
\hepph & \hepth & `(inverse) magnetic catalysis in (2+1)-dimensional gauge theories from holographic models' \\
\hepph & \heplat & `combining infrared and low-temperature asymptotes in yang-mills theories' \\
\hepth & \mathph & `a generalized scaling function for ads/cft' \\
\heplat & \mathph & `green's functions from quantum cluster algorithms'\\
\hepth & \grqc & `quasiparticle universes in bose-einstein condensates'\\
\hepph & \hepth & `on axionic dark matter in type iia string theory' \\
\mathph & \grqc & `fluids in weyl geometries' \\
\heplat & \hepph & `vacuum stability and the higgs boson'\\
\hepth & \heplat & `renormalization in coulomb-gauge qcd within the lagrangian formalism'\\
\hline
\end{tabular}
\caption{{\sf Examples of titles that were mis-classified by the support vector classifier. \label{t:misclass}}}
\end{footnotesize}
\end{table}

Regular contributors to these sections of the \arXiv\ may be curious to see, amongst those which have been mis-classified, what sorts of titles they are. We exhibit a few of the mis-classified cases in Table~\ref{t:misclass}. Is is not hard to see why these titles were mis-matched. For example, the words `cosmological and black-hole' have made the first title more like \grqc, and `coulomb-gauge qcd' in the last title more like \heplat.

\subsection{Cross-Checking Results}

We now make a few remarks about the robustness of our methodology.
First, it is important that we established a {\it single} word vector space $V$. 
As a sanity check, we established {\it separate} vector-spaces, one for each section, and used the SVM to classify, and obtained, rather trivially, almost the identity matrix for $\cM$. The typical false-positive rate was of order 0.3\%. This means that the vector spaces created by the vector embeddings for the five sections are highly disjoint, with almost no overlap in the embedding space of $\IR^{400}$, despite the similar vocabulary employed. Of course, establishing different vector-spaces a priori is useless for a classification problem since categorizing into different section is precisely the problem to be solved.

Next, we can check that titles {\it within} the same section can be consistently classified.
To test this, we train 20,000 titles from \hepth\ into a single vector space, but separate into two random, non-overlapping, groups, labelled as 1 and 2.
Repeating the same procedure as above, we obtained a $2 \times 2$ confusion matrix 
\begin{equation} \cM = \left(
\begin{array}{cc}
 54.8 & 45.3 \\
 54.7 & 45.0 \\
\end{array}
\right)\, ,
\label{thvsth}
\end{equation}
where we will report percentages, as opposed to raw numbers, in the remaining confusion matrices.
The fact that this matrix is almost perfectly divided into the four blocks is very reassuring.
It strongly shows that titles coming from the same section, when translated into word-vectors, are indistinguishable for the SVM.

Finally, we comment on the importance of the \wordvec\ neural network. One might imagine that in the classification algorithm one could bypass \wordvec\ completely and use, instead, something much more straight-forward. Consider the following alternative mechanism. The set of all titles in each of the five \arXiv\ sections is already a labelled data set. One could simply establish a vocabulary of words for each section and lexicographically order them. Then, each title in a section is a list of words which can be labelled by an integer vector, where each word is replaced by its position in the vocabulary. This is essentially replacing the \wordvec -generated vector embedding with a trivial (one-hot) embedding (in a single vector space). This collection of vectors, together with the section label, can then form the training data for a standard application of supervised machine learning.

We performed this straight-forward exercise, using our same SVM architecture to attempt to classify titles, with a validation set of 50,000~titles. 
We find that the result of the predictor is rather random across the sections, as shown by the percentage confusion matrix
\begin{equation}\label{onehot}
\begin{array}{c|ccccc}
$\backslashbox{Actual}{\wordvec + SVM}$ & 1 & 2 & 3 & 4 & 5 \\ \hline
 1 & 42. & 3.7 & 20. & 27. & 6.9 \\
 2 & 36. & 5.2 & 28. & 23. & 8.3 \\
 3 & 36. & 4.3 & 32. & 20. & 8.5 \\
 4 & 30. & 3.3 & 17. & 20. & 30. \\
 5 & 37. & 4.2 & 29. & 22. & 8.1 \\
\end{array}
\end{equation}
which is far from diagonal (and interestingly, mostly being mis-classified as \hepth). We conclude that simply knowing the typical vocabulary of an author in \hepth, versus one in, say, \grqc, is insufficient to sort papers into \arXiv\ sections. In contrast \wordvec\ not only knows the words, it also has learnt the context of words by knowing their nearby neighbours.
Knowing this context is crucial to our analysis, i.e., SVM without \wordvec\ is ineffectual.


\subsection{Beyond Physics}
To further re-assure ourselves of the validity and power of \wordvec\ in conjunction with SVM, we can perform a number of interesting cross-checks.
Suppose we took all titles from a completely different section of the \arXiv\ which, {\it a priori}, should not be related at all to any of our five physics sections. 
What about beyond the \arXiv? How well does the neural network perform?
To answer, let us take the following titles:
\begin{description}
\item[(1-5)] as thus far studied: \hepth, \hepph, \heplat, \grqc, \mathph;
\item[(6)] \condmat: This is the condensed matter physics section. Beginning in April of 1992, it consists of research related to material science, superconductivity, statistical mechanics, etc.; there are many papers cross-listed between \hepth\ and \condmat, especially after the the AdS/CMT correspondence.
\item[(7)] \qfin: The quantitative finance section, beginning in December of 2008, is one of the newest sections to the \arXiv.
\item[(8)] \stat: Another newcomer to the \arXiv, the statistics section began in April of 2007, and consists of such topical fields as machine learning.
\item[(9)] \qbio: Receiving contributions since September of 2003, quantitative biology is an important section of the \arXiv\ consisting of the latest research in mathematical biology and bio-physics.
\item[(10)] \india: This is our control sample.   The {\em Times of India} headlines \cite{headline} are available online, a compilation of 2.7 million news headlines published by {\em Times of India} from 2001 to 2017, from which we extract 20,000 random samples.
These present a reasonably good analogue to \arXiv\ titles in terms of length and syntax. 
\item[(11)] \vhep: An alternative to \arXiv\ is the so-called {\sf viXra} (which is \arXiv\ spelt backwards) \cite{vixra}. Set up by independent physicist Philip Gibbs as an alternative to \arXiv\ in 2007, it aims to cover topics across the whole scientific community, accepting submissions without requiring authors to have an academic affiliation, and does not have the quality control which \arXiv\ has.
In fact, it typically consists of papers rejected by \arXiv\ and has not been accepted by any of the major peer-reviewed journals.\footnote{To give an example, one single contributor to {\sf viXra} has submitted 2572~manuscripts since December of 2011 -- a publication rate that exceeds one per calendar day. It seems clear that such output cannot be compatible with the normal standards of rigor and novelty that is standard in the academic community.} So far {\sf viXra} has around 20,000 total titles across all disciplines in science, though dominated by physics and mathematics.\\

\vhep\ is the high energy physics section of {\sf viXra}, and up to the end of 2017, consists of  1233 titles;
\item[(12)] \vqgst: Likewise, this is the Quantum Gravity and String Theory section of {\sf viXra}, and up to the end of 2017, consists of 1494 titles.
Both \vhep\ and \vqgst\ are relatively small in size but we will nevertheless study them as a curiosity.
\end{description}
We remark that we have specifically included the above three sections of (7), (8) and~(9) because many practitioners of the high-energy theory and mathematical physics community,  who once posted to sections (1)-(5), have at various points of their careers switched to these three fields. It would be interesting to see whether any linguistic idiosyncrasies are carried over. 

Now, it is obvious that both {\sf viXra} sections have, unfortunately, much smaller sample size than the rest, so while we will include them when constructing the conglomerate word-vector embedding by \wordvec, it is sensible to single them out for the classification stage. We find that \textbf{no} title from categories (1-10) is mis-classified into either of the two sections \vhep\ and \vqgst, and these two sections are almost randomly classified into the others. To some extent, this represents the disproportionately small size of the two {\sf viXra} sections. But it also suggests that these particular sections do not even have a self-consistent linguistic structure. The classification probabilities for the two {\sf viXra} sections into other categories are
\begin{equation}\label{vixra}
\begin{array}{c|cccccccccc}
$\backslashbox{Actual}{NN}$
    & 1 & 2 & 3 & 4 & 5 & 6 & 7 & 8 & 9 & 10  \\ \hline
 \vhep\ & 11.5 & 47.4 & 6.8 & 13. & 11. & 4.5 & 0.2 & 0.3 & 2.2 & 3.1 \\
 \vqgst & 13.3 & 14.5 & 1.5 & 54. & 8.4 & 1.8 & 0.1 & 1.1 & 2.8 & 3. \\
\end{array}
\end{equation}
suggesting that titles in \vhep\ are most like titles in \hepph\, while those in \vqgst\ are most like \grqc. If one were to peruse the titles in {\sf viXra}, this correspondence would become readily apparent.


For the remaining ten sections, we find the percentage confusion matrix to be:
\comment{
{{0, 1, 2, 3, 4, 5, 6, 7, 8, 9, 10, 11, 12}, {1, 39., 5.6, 9.4, 25., 
  15.8, 4.13, 0.0179, 0.402, 0.52, 0.17, 0, 0}, {2, 7.9, 61., 14.9, 
  8.6, 2.87, 3.4, 0.0189, 0.83, 0.35, 0.425, 0, 0}, {3, 6.2, 9.2, 72.,
   1.19, 4.6, 4.51, 0.026, 1.15, 0.58, 0.242, 0, 0}, {4, 11.5, 3.7, 
  0.71, 72., 8.5, 1.83, 0.0268, 0.9, 0.8, 0.384, 0, 0}, {5, 11.3, 
  0.87, 2.87, 7.1, 63., 9.8, 0.088, 2.58, 2.21, 0.167, 0, 0}, {6, 2.4,
   1.19, 4.07, 1.43, 8.5, 73., 0.469, 1.65, 6.7, 0.146, 0, 0}, {7, 
  0.61, 0.476, 0.265, 0.71, 5.9, 1.43, 60., 18.3, 11.4, 1.38, 0, 
  0}, {8, 0.249, 0.487, 0.476, 0.63, 5.4, 0.476, 1.7, 81., 9.2, 0.7, 
  0, 0}, {9, 0.447, 1.04, 0.51, 0.71, 3.92, 7.3, 0.434, 12.6, 72., 
  0.87, 0, 0}, {10, 0.52, 1.51, 0.116, 1.28, 0.52, 0.76, 0.465, 0.32, 
  2.15, 92., 0, 0}, {11, 11.5, 47.4, 6.8, 13., 11., 4.45, 0.171, 
  0.342, 2.23, 3.08, 0, 0}, {12, 13.3, 14.5, 1.49, 54., 8.4, 1.79, 
  0.149, 1.05, 2.84, 2.99, 0, 0}}
  }
  
\begin{equation}\label{CMbig}
\begin{array}{c|cccccccccc}
$\backslashbox{Actual}{NN+SVM}$
    & 1 & 2 & 3 & 4 & 5 & 6 & 7 & 8 & 9 & 10 \\ \hline
 1 & 39. & 5.6 & 9.4 & 25. & 15.8 & 4.1 & 0.02 & 0.4 & 0.5 & 0.2 \\
 2 & 7.9 & 61. & 14.9 & 8.6 & 2.9 & 3.4 & 0.02 & 0.8 & 0.4 & 0.4 \\
 3 & 6.2 & 9.2 & 72. & 1.2 & 4.6 & 4.5 & 0.03 & 1.2 & 0.6 & 0.2 \\
 4 & 11.5 & 3.7 & 0.7 & 72. & 8.5 & 1.8 & 0.03 & 0.9 & 0.8 & 0.4 \\
 5 & 11.3 & 0.9 & 2.9 & 7.1 & 63. & 9.8 & 0.09 & 2.6 & 2.2 & 0.2 \\
 6 & 2.4 & 1.2 & 4.1 & 1.4 & 8.5 & 73. & 0.5 & 1.7 & 6.7 & 0.1 \\
 7 & 0.6 & 0.5 & 0.3 & 0.7 & 5.9 & 1.4 & 60. & 18.3 & 11.4 & 1.4 \\
 8 & 0.3 & 0.5 & 0.5 & 0.6 & 5.4 & 0.5 & 1.7 & 81. & 9.2 & 0.7 \\
 9 & 0.4 & 1. & 0.5 & 0.7 & 3.9 & 7.3 & 0.4 & 12.6 & 72. & 0.9 \\
 10 & 0.5 & 1.5 & 0.1 & 1.3 & 0.5 & 0.8 & 0.5 & 0.3 & 2.2 & 92. \\
\end{array}
\end{equation}
where we present the results in percentage terms because some of the title sets studied have different numbers of samples.
Reassuringly, $\cM$ is very much diagonal. Even more significant is the fact that the greatest percentage correct (92\%) is (10), corresponding to the newspaper headline: the syntax and word-choice of the world of journalism is indeed very different from that of science.
The next higher score is 81\%, for (8), \stat, while, interestingly, \qfin\ is not that markedly different from the physics sections.

To be more quantitative, we can reduce this data set to a number of binary ``X'' versus ``not-X'' classifications, and reduce the confusion matrix accordingly. For example, singling out only row/column ten, {\em Times of India} headlines, we find a confusion matrix of
\begin{equation}\label{times}
\begin{array}{c|cc}
$\backslashbox{Actual}{\wordvec + SVM}$
    & {\sf Times} & {\sf not-Times} \\ \hline
 {\sf Times} & 3176 & 263 \\
 {\sf not-Times} & 353 & 85040 \\
\end{array}\, ,
\end{equation}
corresponding to an overall accuracy of 99.3\% for this particular classification. That scholarly publication titles can be separated from newspaper headlines with less than 0.7\% inaccuracy may seem trivial, but this level of fidelity of \wordvec\ is generally not seen in canonical classification challenges.

It is also instructive to see how well our neural network was able to distinguish natural science (physics + biology) from everything else (statistics, quantitative finance, and newspaper headlines), and how well it can distinguish high energy physics from everything else. In the latter case, we are asking for the separation between titles in the five high energy physics (HEP) sections and those in the condensed-matter section \condmat. The separation of natural science gives a confusion matrix of
\begin{equation}\label{natsci}
\begin{array}{c|cc}
$\backslashbox{Actual}{\wordvec + SVM}$
    & {\sf Natural Science} & {\sf not-Natural Science} \\ \hline
 {\sf Natural Science} & 69223 & 2137 \\
 {\sf not-Natural Science} & 2584 & 13733 \\
\end{array}\, ,
\end{equation}
which corresponds to an accuracy of 94.6\%, whereas the separation between HEP sections and non-HEP sections is only slightly less accurate
\begin{equation}\label{hep}
\begin{array}{c|cc}
$\backslashbox{Actual}{\wordvec + SVM}$
    & {\sf HEP} & {\sf not-HEP} \\ \hline
 {\sf HEP} & 50882 & 3825 \\
 {\sf not-HEP} & 3325 & 30799 \\
\end{array}\, ,
\end{equation}
corresponding to an accuracy of 92.0\%.

\comment{
We see that {\it none} of the headlines have been mistaken to be an \arXiv\ title (the final row of zeros), which is reassuring.
Of the entries in the final column -- actual \arXiv\ titles that were mis-classified as newspaper headlines -- they include some obviously whimsical ones such as 
`27/32' (\arXiv:0906.0965 ) and `Juggler's exclusion process', (\arXiv:1104.3397); particularly short titles such as `Metaplectic Ice' (\arXiv:1009.1741) and
`kline foams' (\arXiv:1001.4781); biographies such as `gregor wentzel' (\arXiv:0809.2102); and titles which are structured as questions such as
`classical/quantum = commutative/noncommutative?' (\arXiv:1204.1858), or with colons such as `quantum geometrodynamics: whence, whither?' (\arXiv:0812.0295).
}

\vspace{1in}

%
%
\section{Conclusion}
\label{sec:conclude}

In this paper we have performed a systematic textual analysis of theoretical particle physics papers using natural language processing and machine learning techniques. Focusing specifically on titles of \hepth\ papers, and then later on the titles of papers in four related sections of the \arXiv, we have demonstrated the ability of a classifying agent, informed by vector word embeddings generated by the continuous bag-of-words neural network model of \wordvec, to accurately classify papers into five \arXiv\ sections, using only paper titles, with an accuracy of just over 65\%. For the slightly easier task of separating high-energy physics titles from those of other scientific pursuits -- even other branches of theoretical physics -- the classification accuracy was 92\%.

%
We have demonstrated that this classification accuracy is not adequately explained by distinctions between the words themselves that are employed by authors in the various sections. Rather, the contextual windows, captured by the vector space embeddings constructed by \wordvec, are clearly important to the ability to distinguish titles. A practitioner of theoretical particle physics, such as the authors, could hardly be expected to achieve such accuracy, which is, of course, why machine learning is so powerful in these sorts of problems. In fact, the use of natural language processing to classify documents is well established in the data analysis community. We would like to suggest the performance in classifying papers in \arXiv\ be used as a benchmark test for classification algorithms generally, as the differences between the sections are subtle, even to long-time contributors to the field.

Along the way to demonstrating the classification ability of \wordvec, we discovered a number of interesting properties of \arXiv\ authors, and \hepth\ specifically. Many of these observations belong to a growing field of meta-analysis in the sciences that has come to be known as {\it scientometrics}~\cite{metrics}. In particular, the contextual analysis revealed strong ties between \hepth\ and the more formal branches of the larger high-energy theory community: \mathph\ and \grqc. The connection between the syntax used by \hepth\ authors had close affinity to \hepph, but not to \heplat, though the latter two sections of \arXiv\ were themselves very similar in syntax and subject content.

As those who have been in the theoretical high energy community for some time can attest, there has long been a perception of a rather wide chasm between ``formal theorists'' and ``phenomenologists'', and that this sociological chasm is born out in the bifurcation of our field in terms of conference/workshop attendance, citation, co-authorship, etc.\footnote{The relatively small string phenomenology community has a negligible effect on this division.} Such compartmentalization of the high energy community can even be seen visually, in a representation of the citation network across various branches of high energy physics~\cite{paperscape}. 

Finally, this work has a contribution to make to the more formal study of vector space word embeddings, and natural language processing more generally. As mentioned earlier in the text, the ``strange geometry'' of vector space word embeddings is an interesting area of study. The tendency of word vectors to cluster in the positive orthant, and the highly conical nature of the assigned vectors, has been studied in other contexts~\cite{Mimno}. Typically these data sets are of a more general nature, like the {\it Times of India} headlines, whose affinity distances were shown in Section~\ref{sec:distance}. For the highly technical and specific papers that form the corpus of \hepth, however, these geometrical properties are even more pronounced. Indeed, we suggest that the degree of conical clustering in a particular word embedding may serve as a marker for the degree of specificity of a particular corpus. Thus, we might expect more clustering in headlines drawn strictly from the finance section of the newspaper, than from newspaper headlines generally. On a more mathematical note, it would be of interest to explore the true dimensionality of our vector space word embedding (undoubtedly far smaller than the 400~dimensional embedding space we have chosen), and to study the properties of the space orthogonal to any particular word vector. We leave such issues to a future work.

Ultimately, it would be of interest to devote more attention to the abstracts of \arXiv\ papers, and eventually to the text of those papers themselves. The decision to restrict ourselves primarily to titles was mainly made on the basis of limited computational resources, though there are technical issues to consider as well. In dealing with abstracts (and eventually whole documents), two methods of attack immediately present themselves:
(a) we could to take {\it each} abstract as a document and then study cross-document word embeddings;
(b) we could take the full list of abstracts as a single document, each sentence being a proper sentence in the English language, separated by the full-stop. It is clear that (b) is more amenable to \wordvec, though it would clearly obscure the contextual differences between one paper and the next if context windows were to cross abstracts. We hope to return to this issue, perhaps employing variants of \wordvec\ such as \texttt{Doc2vec}~\cite{doc2vec}, which takes not a list of list of words, but rather a triple layer of a list of list of list of words, and which is obviously more suited for method (a).

\section*{Acknowledgements}
We are indebted to Paul Ginsparg for his many suggestions and helpful comments.
YHH would like to thank the Science and Technology Facilities Council, UK, for grant ST/J00037X/1, 
the Chinese Ministry of Education, for a Chang-Jiang Chair Professorship at NanKai University and the City of Tian-Jin for a Qian-Ren Scholarship.
VJ is supported by the South African Research Chairs Initiative of the DST/NRF.

\newpage

\appendix

\section{From Raw Titles to Cleaned Titles}
\label{ap:clean}

\begin{figure}[t]
\centerline{
\includegraphics[trim=15mm 0mm 0mm 0mm, clip, width=6.5in]{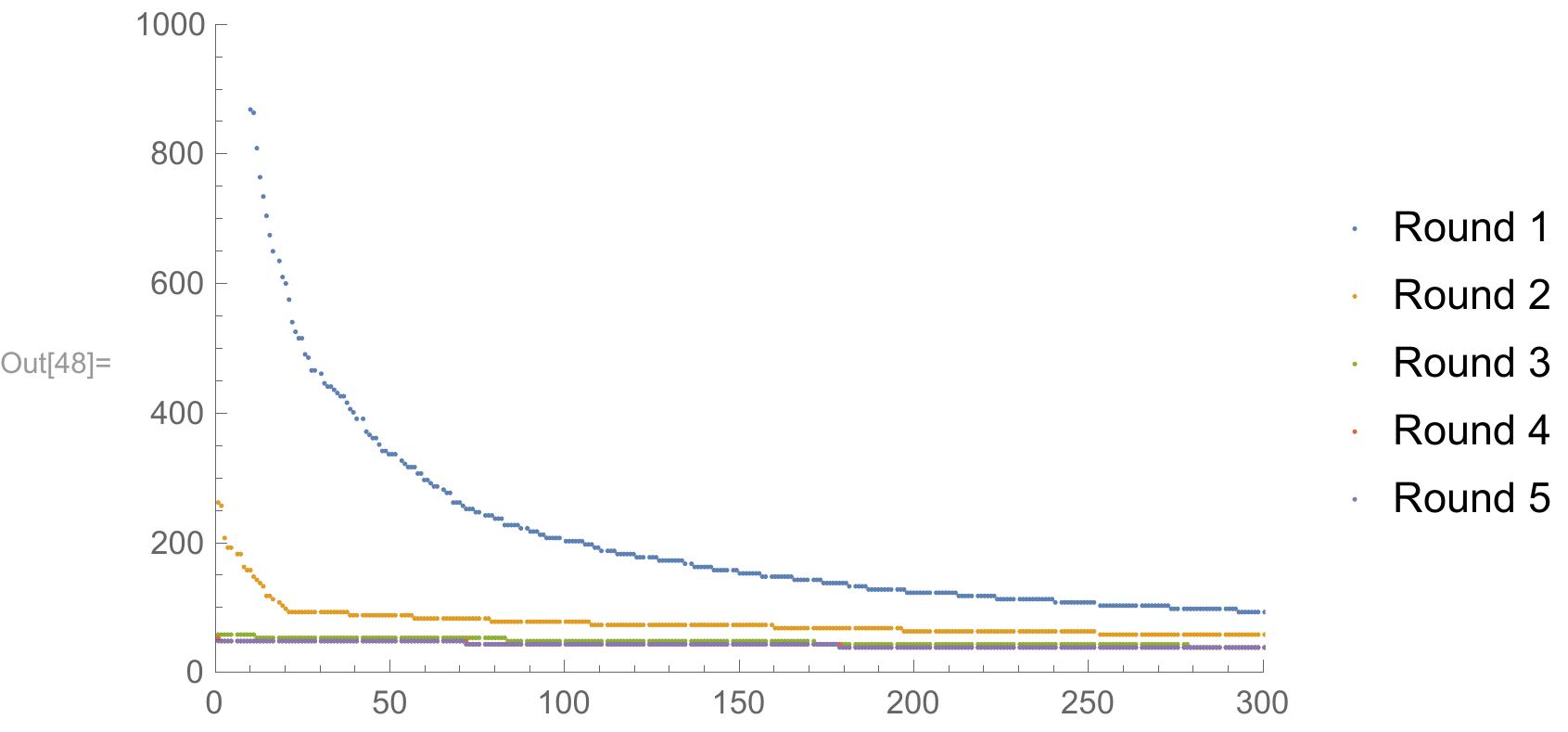}
}
\caption{{\sf {\small
Histogram of frequency of appearance of 2-grams after 5 rounds of hyphenating the 2-grams with frequencies exceeding 50.
We see that after 2 iterations the distribution is essentially flat, signifying that words which should be hyphenated have been.
\label{f:replace-hepth}}}}
\end{figure}

In this Appendix, we describe the process and consequences of cleaning.
We work with $120,249$ titles from \hepth, with a total of $996,485$ words, of which $48,984$ are unique.
First, we convert the text to lower case and remove punctuation.
Next, after replacing plurals, foreign spellings and some technical acronyms (such as ``ramond ramond'' $\to$ ``rr''), we are left with $701,794$ words of which $32,693$ are unique.

Now, we can let the computer find the most common $2$-grams (words which appear adjacent to each other).
The top hits are ``gauge theory'', ``field theory'', ``scalar field'', etc.
These terms are collective nouns.
They clearly need to be hyphenated and considered as single words.
Plotting the frequency of the top $300$ $2$-grams, one can see that there is a tailing off at around $50$ or so (\textit{cf.}~Figure \ref{f:replace-hepth}).
This means that we should simply hyphenate $2$-grams that appear more than $50$ times.

We repeat this process, hyphenating on each iteration those $2$-grams that appear with a frequency exceeding $50$.
This can concatenate strings to form $3$-grams, such as the word ``quantum-field-theory''. 
Figure~\ref{f:replace-hepth} shows that the histogram becomes essentially flat after two iterations.
This is quite interesting as it signifies that technical words which should be hyphenated can be detected automatically.
We will use the flatness of the curve to stop the automatic replacements after two rounds of hyphenation.
For reference, we record the most common ten words at each round of replacements, as well as the number of times that they appear.

\begin{tabularx}{\textwidth}{|l|X|}\hline
 \text{Round 1} & $\{\{\{$gauge, theory$\}$, 3232$\}$, $\{\{$field, theory$\}$, 2937$\}$,
   $\{\{$string, theory$\}$, 2025$\}$, $\{\{$quantum, gravity$\}$, 1552$\}$, $\{\{$yang-mills,
   theory$\}$, 1207$\}$, $\{\{$scalar, field$\}$, 1174$\}$, $\{\{$quantum, mechanics$\}$, 1117$\}$,
   $\{\{$dark, energy$\}$, 1084$\}$, $\{\{$matrix, model$\}$, 1040$\}$, $\{\{$cosmological,
   constant$\}$, 869$\}\}$ \\\hline
 \text{Round 2} & $\{\{\{$string-field, theory$\}$, 260$\}$, $\{\{$susy-gauge, theory$\}$,
   255$\}$, $\{\{$loop-quantum, gravity$\}$, 209$\}$, $\{\{$scalar-field, theory$\}$, 191$\}$,
   $\{\{$chiral-symmetry, breaking$\}$, 191$\}$, $\{\{$n=4-sym, theory$\}$, 182$\}$,
   $\{\{$lattice-gauge, theory$\}$, 182$\}$, $\{\{$topological-field, theory$\}$, 162$\}$,
   $\{\{$noncommutative-gauge, theory$\}$, 158$\}$, $\{\{$susy-quantum, mechanics$\}$, 155$\}\}$ \\\hline
 \text{Round 3} & {$\{\{\{$pair, creation$\}$, 56$\}$, $\{\{$cosmic, censorship$\}$, 56$\}$,
   $\{\{$holographic, principle$\}$, 56$\}$, $\{\{$three, dimensional$\}$, 56$\}$, $\{\{$logarithmic,
   correction$\}$, 56$\}$, $\{\{$dimensional, regularization$\}$, 56$\}$, $\{\{$nonlinear, susy$\}$,
   56$\}$, $\{\{$susy, standard-model$\}$, 56$\}$, $\{\{$stochastic, quantization$\}$, 56$\}$,
   $\{\{$density, perturbation$\}$, 56$\}\}$} \\\hline
 \text{Round 4} & {$\{\{\{$heavy-ion, collisions$\}$, 52$\}$, $\{\{$new, massive-gravity$\}$,
   49$\}$, $\{\{$conformal, algebra$\}$, 49$\}$, $\{\{$warped, compactification$\}$, 49$\}$,
   $\{\{$group, manifold$\}$, 49$\}$, $\{\{$weak, gravity$\}$, 49$\}$, $\{\{$stress-energy, tensor$\}$,
   49$\}$, $\{\{$elementary, particle$\}$, 49$\}$, $\{\{$bose, gas$\}$, 49$\}$, $\{\{$new, results$\}$,
   49$\}\}$} \\\hline
 \text{Round 5} & {$\{\{\{$new, massive-gravity$\}$, 49$\}$, $\{\{$conformal, algebra$\}$, 49$\}$,
   $\{\{$warped, compactification$\}$, 49$\}$, $\{\{$group, manifold$\}$, 49$\}$, $\{\{$weak,
   gravity$\}$, 49$\}$, $\{\{$stress-energy, tensor$\}$, 49$\}$, $\{\{$elementary, particle$\}$, 49$\}$,
   $\{\{$bose, gas$\}$, 49$\}$, $\{\{$new, results$\}$, 49$\}$, $\{\{$brownian, motion$\}$, 49$\}\}$} \\\hline
\end{tabularx}

\section{Higher $n$-Grams across the Sections}
\label{ap:data}
In this appendix, we will present some statistic of the higher $n$-grams for the various \arXiv\ sections, extending the 1-gram (words) and 2-gram analyses in Sections \ref{s:monogram} and \ref{s:bigram}.
The $15$ most common $3$-grams in \hepth\ titles are:
\begin{tabularx}{\textwidth}{l|X}
\text{Raw}&
\{\{\text{quantum},\text{field},\text{theory}\},826\}, \{\{\text{a},\text{note},\text{on}\},620\}, \{\{\text{conformal},\text{field},\text{theory}\},477\}, \{\{\text{in},\text{string},\text{theory}\},475\}, \{\{\text{string},\text{field},\text{theory}\},399\}, \{\{\text{the},\text{cosmological},\text{constant}\},395\}, \{\{\text{at},\text{finite},\text{temperature}\},390\}, \{\{\text{the},\text{presence},\text{of}\},353\}, \{\{\text{in},\text{the},\text{presence}\},346\}, \{\{\text{the},\text{standard},\text{model}\},329\}, \{\{\text{conformal},\text{field},\text{theories}\},324\}, \{\{\text{field},\text{theory},\text{and}\},284\}, \{\{\text{in},\text{de},\text{sitter}\},281\}, \{\{\text{approach},\text{to},\text{the}\},273\}, \{\{\text{corrections},\text{to},\text{the}\},263\}
\\\hline
\text{Cleaned}&
\{\{\text{weak},\text{gravity},\text{conjecture}\},41\}, \{\{\text{causal},\text{dynamical},\text{triangulations}\},37\}, \{\{\text{strong},\text{cp},\text{problem}\},30\}, \{\{\text{string},\text{gas},\text{cosmology}\},29\}, \{\{\text{van},\text{der},\text{waals}\},28\}, \{\{\text{ads$\_$5},\text{times},\text{s${}^{\wedge}$5}\},25\}, \{\{\text{shape},\text{invariant},\text{potential}\},24\}, \{\{\text{chern-simons},\text{matter},\text{theory}\},24\}, \{\{\text{ads/cft},\text{integrability},\text{chapter}\},23\}, \{\{\text{review},\text{ads/cft},\text{integrability,}\},23\}, \{\{\text{type},0,\text{string-theory}\},23\}, \{\{\text{closed},\text{timelike},\text{curves}\},22\}, \{\{\text{hard},\text{thermal},\text{loop}\},22\}, \{\{\text{lowest},\text{landau},\text{level}\},22\}, \{\{\text{varying},\text{speed},\text{light}\},21\}
\\
\end{tabularx}

Once again, more information is to be found in the cleaned $3$-grams.
We conclude here as well that the most common terms such as ``weak gravity conjecture'', ``causal dynamical triangulations'', or ``the strong CP problem'' should be collective nouns with more stringent cleaning. 

Finally, the $15$ most common $4$-grams in \hepth\ titles are:
\begin{tabularx}{\textwidth}{l|X}
\text{Raw}&
\{\{\text{in},\text{the},\text{presence},\text{of}\},345\}, \{\{\text{in},\text{quantum},\text{field},\text{theory}\},201\}, \{\{\text{a},\text{note},\text{on},\text{the}\},181\}, \{\{\text{and},\text{the},\text{cosmological},\text{constant}\},145\}, \{\{\text{the},\text{cosmological},\text{constant},\text{problem}\},120\}, \{\{\text{in},\text{de},\text{sitter},\text{space}\},111\}, \{\{\text{of},\text{the},\text{standard},\text{model}\},94\}, \{\{\text{open},\text{string},\text{field},\text{theory}\},89\}, \{\{\text{the},\text{presence},\text{of},\text{a}\},89\}, \{\{\text{in},\text{conformal},\text{field},\text{theory}\},82\}, \{\{\text{chiral},\text{symmetry},\text{breaking},\text{in}\},81\}, \{\{\text{in},\text{a},\text{magnetic},\text{field}\},81\}, \{\{\text{of},\text{the},\text{cosmological},\text{constant}\},80\}, \{\{\text{effective},\text{field},\text{theory},\text{of}\},80\}, \{\{\text{the},\text{moduli},\text{space},\text{of}\},77\}
\\\hline
\text{Cleaned}&
\{\{\text{review},\text{ads/cft},\text{integrability},\text{chapter}\},23\}, \{\{\text{solution},\text{strong},\text{cp},\text{problem}\},12\}, \{\{\text{chiral},\text{de},\text{rham},\text{complex}\},9\}, \{\{\text{superstring},\text{derived},\text{standard-like},\text{model}\},8\}, \{\{\text{bundle},\text{formulation},\text{nonrelativistic-quantum},\text{mechanics}\},7\}, \{\{\text{fibre},\text{bundle},\text{formulation},\text{nonrelativistic-quantum}\},7\}, \{\{\text{radiatively},\text{induced},\text{lorentz},\text{cpt-violation}\},7\}, \{\{\text{vortex},\text{model},\text{ir},\text{sector}\},7\}, \{\{\text{center},\text{vortex},\text{model},\text{ir}\},7\}, \{\{\text{ultra},\text{high-energy},\text{cosmic},\text{rays}\},7\}, \{\{\text{radiation},\text{d-dimensional},\text{collision},\text{shock-wave}\},7\}, \{\{\text{5d},\text{standing},\text{wave},\text{braneworld}\},7\}, \{\{\text{logarithmic},\text{tensor},\text{category},\text{theory,}\},7\}, \{\{\text{five-dimensional},\text{tangent},\text{vector},\text{spacetime}\},7\}, \{\{\text{shear-viscosity},\text{entropy},\text{density},\text{ratio}\},7\}\\
\end{tabularx}

The most common cleaned $4$-gram is a reference to the Beisert, \textit{et al.}\ review on the integrable structure of ${\cal N}=4$ super-Yang--Mills theory~\cite{beisert}.
While again certain $4$-grams obviously point to ``a solution to the strong CP problem'' or ``the chiral de Rham complex'', the prevalence of these terms are sufficiently scarce in the database that finding these collective terms automatically is somewhat difficult.
What is striking as a member of the \hepth\ community is that these collective nouns tie in nicely to an expert's conception of what it is that \hepth\ people do. 


In analogy to Figure~\ref{f:cloudhepth}, we consider in Table~\ref{f:cloud_all} word clouds based on cleaned data for the subjects \hepph, \heplat, \grqc, and \mathph.
An expert can associate any of these word clouds with the label of the corresponding subject area.
Thus, it is not entirely surprising that machine learning algorithms can discriminate titles between these areas as well.

A paper may be of interest to readers of the \arXiv\ in more than one subject area and indeed, can reasonably be posted in either of \hepth\ or \hepph, for example.
Cross-listing on the \arXiv\ enlarges the readership of a paper and enables the expression of a diversity of scientific interests.
For these reasons, there is a significant overlap between the terms that appear in the word clouds.
The emphasis of certain themes, however, renders the identification unambiguous.

\begin{table}[t!!!!]
\begin{tabular}{|c|c|c|c||c|c|c|c|}\hline
& \rotatebox{90}{\# Papers} & \rotatebox{90}{\# Unique Words} & Word Cloud &
& \rotatebox{90}{\# Papers} & \rotatebox{90}{\# Unique Words} & Word Cloud
\\ \hline \hline
\rotatebox{90}{{\bf \hepph}} & \rotatebox{90}{133,346} & \rotatebox{90}{46,011} &
\includegraphics[trim=0mm 0mm 0mm 0mm, clip, width=2in]{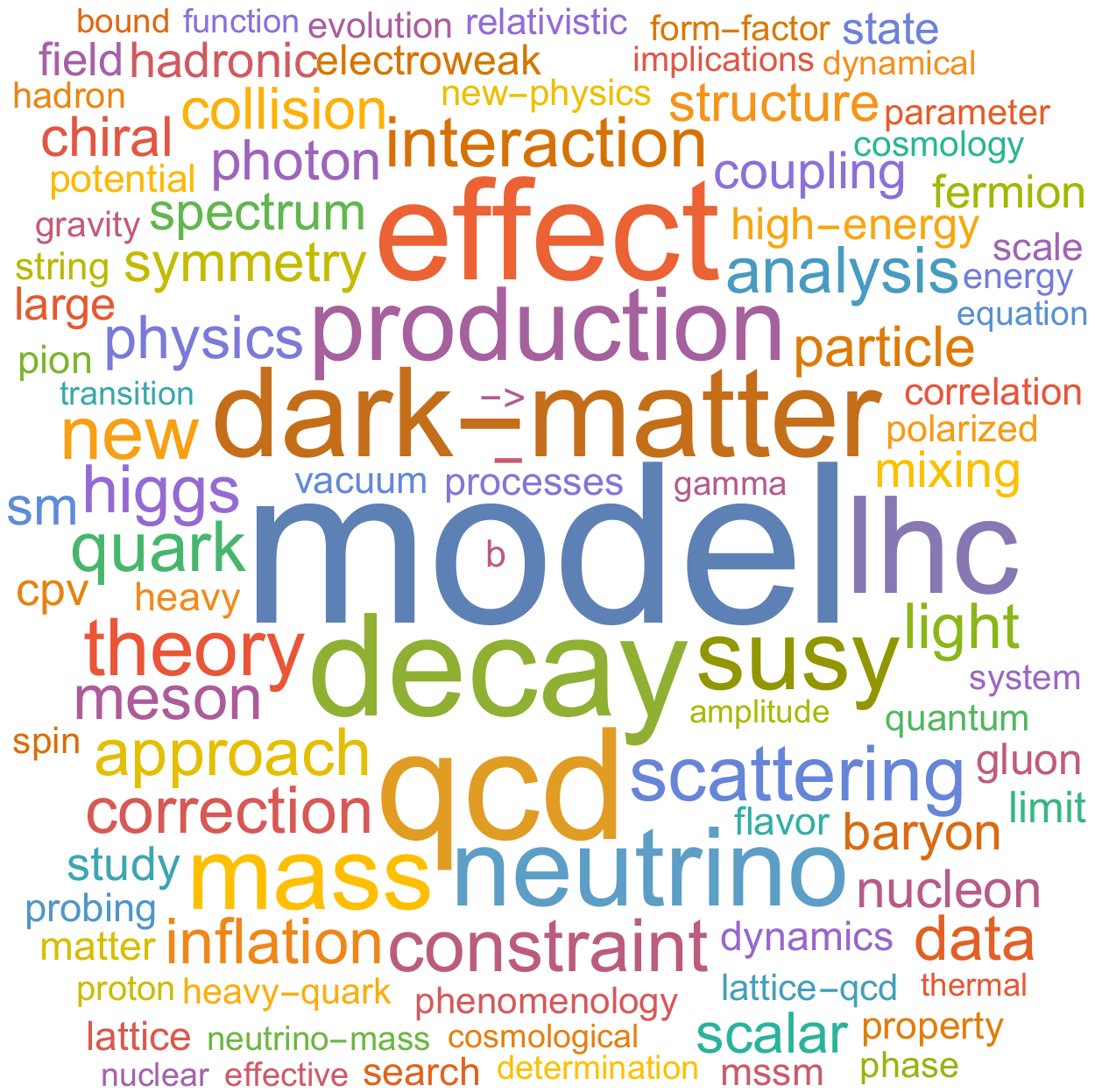}
&
\rotatebox{90}{{\bf \heplat}} & \rotatebox{90}{21,123} & \rotatebox{90}{10,639} &
\includegraphics[trim=0mm 0mm 0mm 0mm, clip, width=2in]{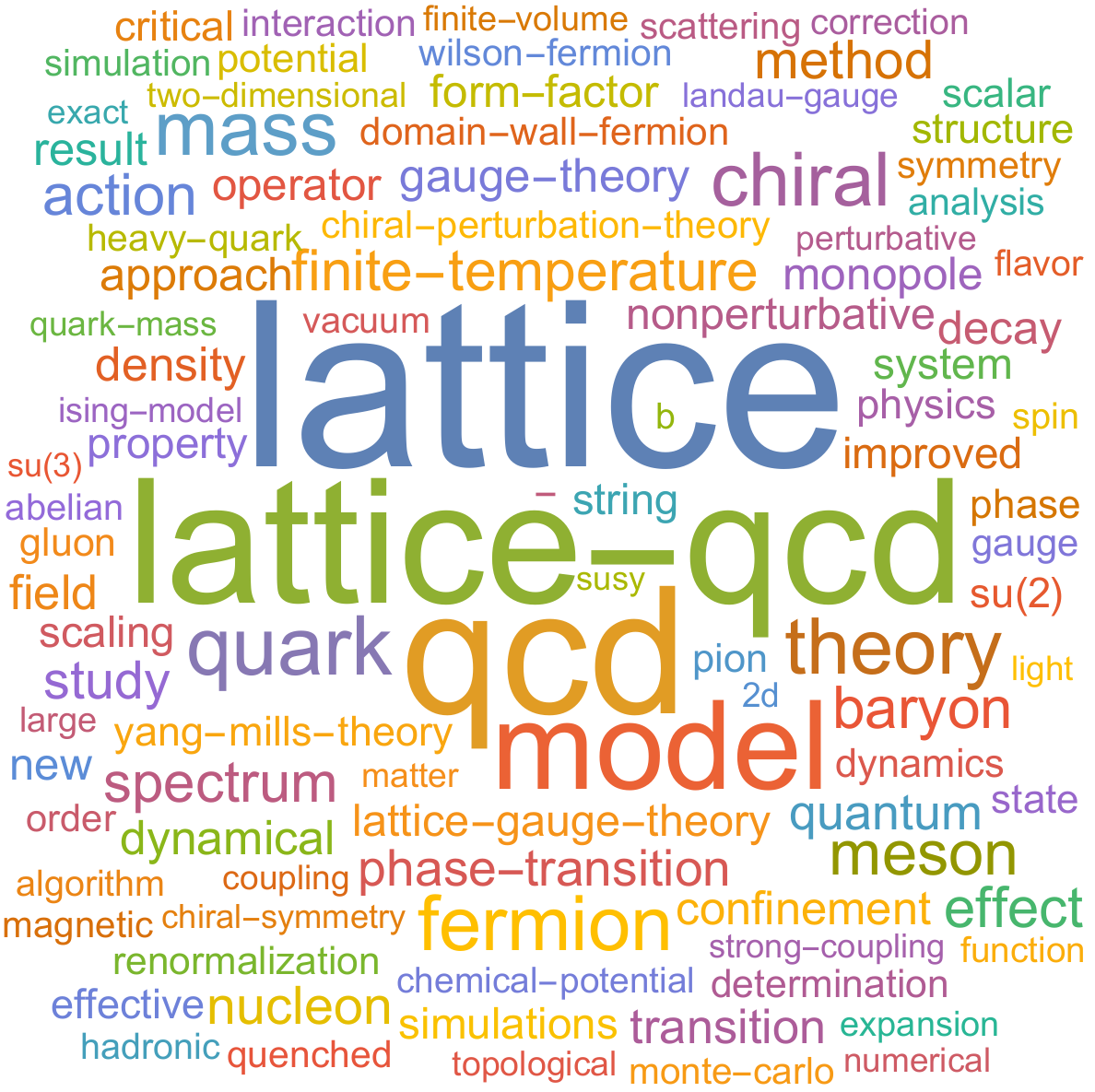}
\\ \hline
\rotatebox{90}{{\bf \grqc}} & \rotatebox{90}{69,386} & \rotatebox{90}{26,222} &
\includegraphics[trim=0mm 0mm 0mm 0mm, clip, width=2in]{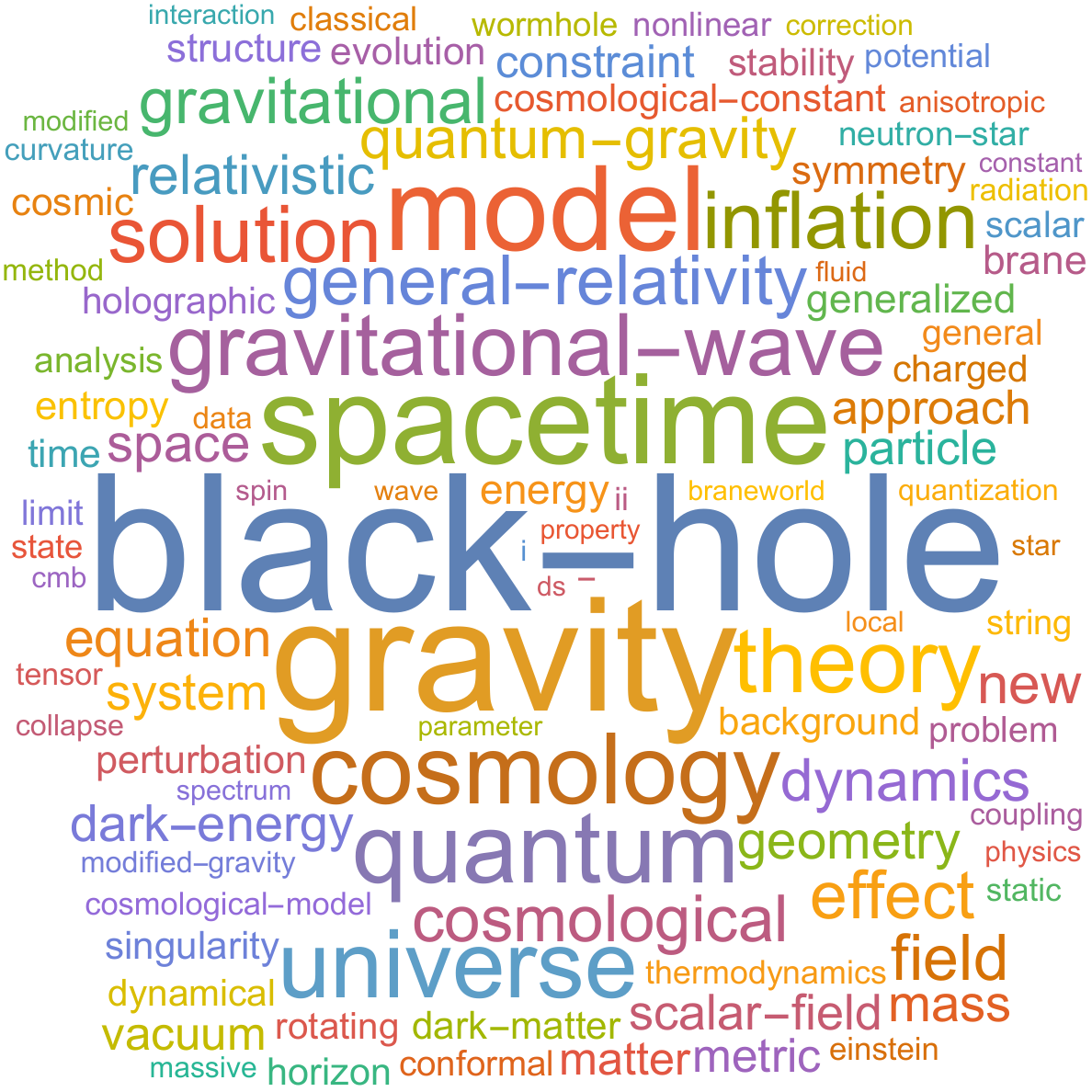}
&
\rotatebox{90}{{\bf \mathph}} & \rotatebox{90}{51,747} & \rotatebox{90}{28,559} &
\includegraphics[trim=0mm 0mm 0mm 0mm, clip, width=2in]{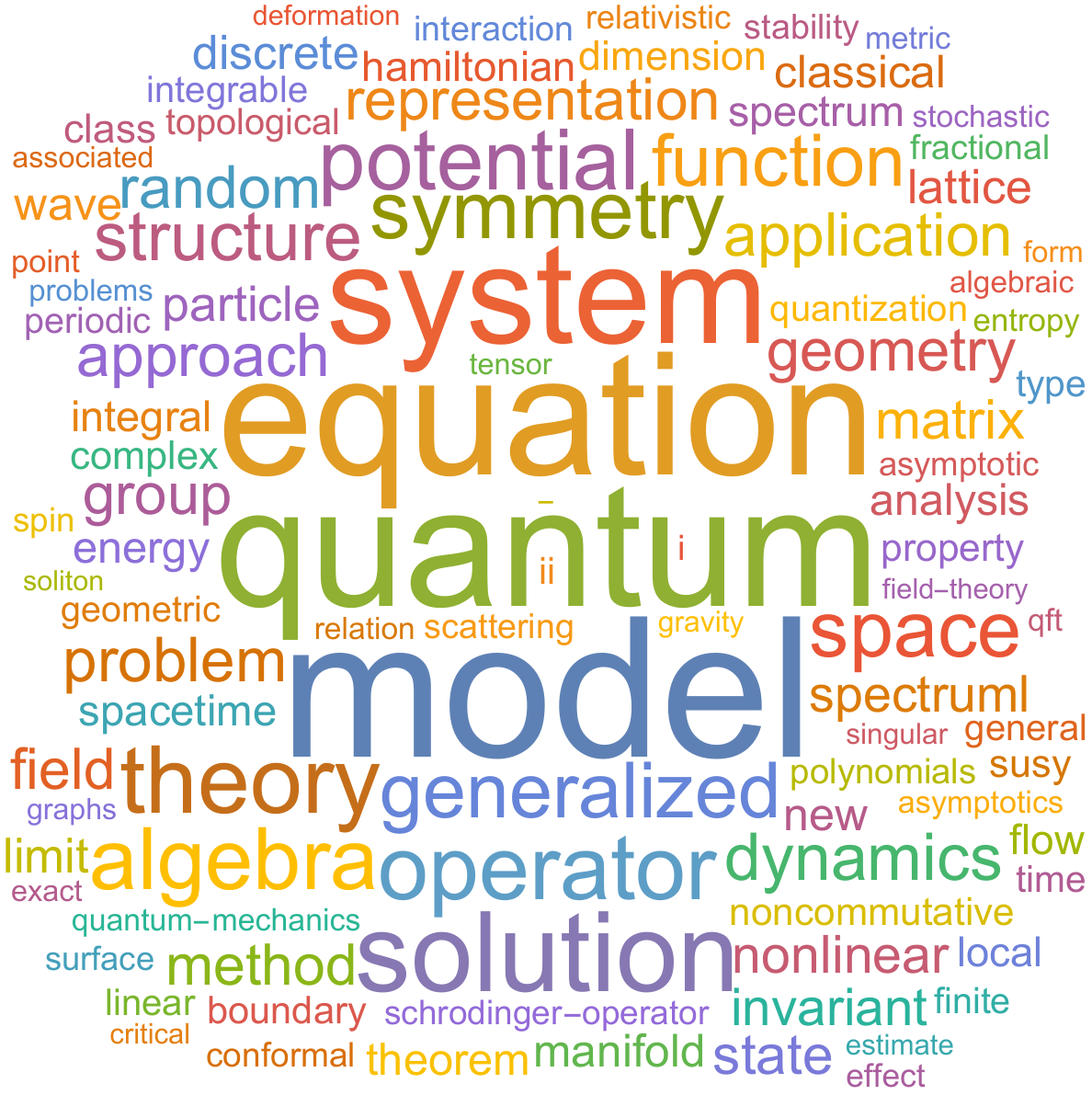}
\\ \hline
\end{tabular}
\caption{{\sf {\small
Some simple statistics of all titles  of the various sections of the high energy \arXiv, from the beginning in 1990 until 2017.
The titles are {\it cleaned} according do the discussions in Section~\ref{s:clean}.
The word cloud is sized according to frequency of relevant words. 
The number of papers and associated number of total unique words were compiled in December 2017.
\label{f:cloud_all}}}}
\end{table}



The most common $15$ $3$-grams in the other \arXiv\ sections based on cleaned data are:
\begin{tabularx}{\textwidth}{l|X}
\text{\hepph}&
\{\text{hidden},\text{local},\text{symmetry}\}, \{\text{hadron},\text{resonance},\text{gas}\}, \{\text{hard},\text{thermal},\text{loop}\}, \{\text{fine},\text{structure},\text{constant}\}, \{\text{higgs},\text{triplet},\text{model}\}, \{\text{t},\text{bar},\text{t}\}, \{\text{hadronic},\text{light-by-light},\text{scattering}\}, \{\text{resonance},\text{gas},\text{model}\}, \{\text{delta},\text{i=1/2},\text{rule}\}, \{\text{inverse},\text{magnetic},\text{catalysis}\}, \{\text{large},\text{momentum},\text{transfer}\}, \{\text{active},\text{galactic},\text{nuclei}\}, \{\text{susy},\text{flavor},\text{problem}\}, \{\text{q},\text{bar},\text{q}\}, \{\text{future},\text{lepton},\text{colliders}\}
\\ \hline
\text{\heplat}&
\{\text{gluon},\text{ghost},\text{propagators}\}, \{\text{electric},\text{dipole},\text{moment}\}, \{\text{hadronic},\text{vacuum},\text{polarization}\}, \{\text{numerical},\text{stochastic},\text{perturbation-theory}\}, \{\text{mass},\text{anomalous},\text{dimension}\}, \{\text{first},\text{order},\text{phase-transition}\}, \{\text{maximum},\text{entropy},\text{method}\}, \{\text{causal},\text{dynamical},\text{triangulations}\}, \{\text{matrix},\text{product},\text{state}\}, \{\text{hadron},\text{resonance},\text{gas}\}, \{\text{chiral},\text{magnetic},\text{effect}\}, \{\text{delta},\text{i=1/2},\text{rule}\}, \{\text{physical},\text{pion},\text{mass}\}, \{\text{neutron},\text{electric},\text{dipole}\}, \{\text{nucleon},\text{axial},\text{charge}\}
\\ \hline
\text{\grqc}&
\{\text{matters},\text{gravity},\text{newsletter}\}, \{\text{einstein},\text{static},\text{universe}\}, \{\text{causal},\text{dynamical},\text{triangulations}\}, \{\text{initial},\text{value},\text{problem}\}, \{\text{modified},\text{newtonian},\text{dynamics}\}, \{\text{topical},\text{group},\text{gravitation}\}, \{\text{van},\text{der},\text{waals}\}, \{\text{eddington-inspired},\text{born-infeld},\text{gravity}\}, \{\text{crossing},\text{phantom},\text{divide}\}, \{\text{baryon},\text{acoustic},\text{oscillation}\}, \{\text{physical},\text{society},\text{volume}\}, \{\text{american},\text{physical},\text{society,}\}, \{\text{newsletter},\text{topical},\text{group}\}, \{\text{gravity},\text{newsletter},\text{topical}\}, \{\text{lunar},\text{laser},\text{ranging}\}
\\ \hline
\text{\mathph}&
\{\text{asymmetric},\text{simple-exclusion},\text{process}\}, \{\text{spectruml},\text{shift},\text{function}\}, \{\text{alternating},\text{sign},\text{matrix}\}, \{\text{mutually},\text{unbiased},\text{bases}\}, \{\text{shape},\text{invariant},\text{potential}\}, \{\text{space},\text{constant},\text{curvature}\}, \{\text{quantum},\text{affine},\text{algebra}\}, \{\text{quantum},\text{dynamical},\text{semigroup}\}, \{\text{density},\text{functional},\text{theory}\}, \{\text{position},\text{dependent},\text{mass}\}, \{\text{inverse-scattering},\text{fixed},\text{energy}\}, \{\text{random},\text{band},\text{matrix}\}, \{\text{random},\text{energy},\text{model}\}, \{\text{asymptotic},\text{iteration},\text{method}\}, \{\text{spin},\text{glass},\text{model}\}
\\
\end{tabularx}

Finally, the $15$ most common $4$-grams in the other \arXiv\ sections based on cleaned data are:

\begin{tabularx}{\textwidth}{l|X}
\text{\hepph}&
\{\text{hadron},\text{resonance},\text{gas},\text{model}\}, \{\text{variation},\text{fine},\text{structure},\text{constant}\}, \{35,\text{kev},\text{x-ray},\text{line}\}, \{\text{au+au},\text{collision},\text{sqrts$\_$nn=200},\text{gev}\}, \{130,\text{gev},\text{gamma-ray},\text{line}\}, \{\text{hadronic},\text{light-by-light},\text{scattering},\text{muon-g-2}\}, \{\text{hadronic},\text{light-by-light},\text{scattering},\text{contribution}\}, \{\text{weak},\text{radiative},\text{hyperon},\text{decay}\}, \{\text{after},\text{lhc},\text{run},1\}, \{\text{nambu},-,\text{jona-lasinio},\text{model}\}, \{\text{variable},\text{flavor},\text{number},\text{scheme}\}, \{\text{large},\text{hadron},\text{electron},\text{collider}\}, \{\text{fermi},\text{large},\text{area},\text{telescope}\}, \{\text{flavor},\text{asymmetry},\text{nucleon},\text{sea}\}, \{\text{mu},\text{-$>$},\text{e},\text{gamma}\}
\\ 
\hline
\text{\heplat}&
\{\text{neutron},\text{electric},\text{dipole},\text{moment}\}, \{\text{ground},\text{state},\text{entropy},\text{potts}\}, \{\text{center-vortex},\text{model},\text{ir},\text{sector}\}, \{\text{hadron},\text{resonance},\text{gas},\text{model}\}, \{\text{i=2},\text{pion},\text{scattering},\text{length}\}, \{\text{gluon},\text{ghost},\text{propagators},\text{landau-gauge}\}, \{\text{higgs},\text{boson},\text{mass},\text{bound}\}, \{\text{international},\text{lattice},\text{data},\text{grid}\}, \{\text{landau-gauge},\text{gluon},\text{ghost},\text{propagators}\}, \{\text{color},\text{confinement},\text{dual},\text{superconductivity}\}, \{\text{state},\text{entropy},\text{potts},\text{antiferromagnets}\}, \{\text{hadronic},\text{contribution},\text{muon},\text{g-2}\}, \{\text{nearly},\text{physical},\text{pion},\text{mass}\}, \{\text{weinberg},-,\text{salam},\text{model}\}, \{\text{kaon},\text{mixing},\text{beyond},\text{sm}\}
\\ \hline
\text{\grqc}&
\{\text{american},\text{physical},\text{society},\text{volume}\}, \{\text{newsletter},\text{topical},\text{group},\text{gravitation}\}, \{\text{gravity},\text{newsletter},\text{topical},\text{group}\}, \{\text{matters},\text{gravity},\text{newsletter},\text{topical}\}, \{\text{group},\text{gravitation},\text{american},\text{physical}\}, \{\text{topical},\text{group},\text{gravitation},\text{american}\}, \{\text{gravitation},\text{american},\text{physical},\text{society,}\}, \{\text{matters},\text{gravity},\text{newsletter},\text{aps}\}, \{\text{innermost},\text{stable},\text{circular},\text{orbit}\}, \{\text{stability},\text{einstein},\text{static},\text{universe}\}, \{\text{newsletter},\text{aps},\text{topical},\text{group}\}, \{\text{gravity},\text{newsletter},\text{aps},\text{topical}\}, \{\text{space},\text{affine},\text{connection},\text{metric}\}, \{\text{instanton},\text{representation},\text{plebanski},\text{gravity}\}, \{\text{laser},\text{astrometric},\text{test},\text{relativity}\}
\\ \hline
\text{\mathph}&
\{\text{mean-field},\text{spin},\text{glass},\text{model}\}, \{\text{long},\text{range},\text{scattering},\text{modified}\}, \{\text{deformation},\text{expression},\text{elements},\text{algebra}\}, \{\text{totally},\text{asymmetric},\text{simple-exclusion},\text{process}\}, \{\text{nonlinear},\text{accelerator},\text{problems},\text{wavelets}\}, \{\text{scattering},\text{modified},\text{wave},\text{operator}\}, \{\text{range},\text{scattering},\text{modified},\text{wave}\}, \{\text{spectruml},\text{parameter},\text{power},\text{series}\}, \{\text{matrix},\text{schrodinger-operator},\text{half},\text{line}\}, \{\text{five-dimensional},\text{tangent},\text{vector},\text{spacetime}\}, \{\text{causal},\text{signal},\text{transmission},\text{quantum-field}\}, \{\text{master},\text{constraint},\text{programme},\text{lqg}\}, \{\text{set},\text{spin},\text{values},\text{cayley-tree}\}, \{\text{uncountable},\text{set},\text{spin},\text{values}\}, \{\text{model},\text{uncountable},\text{set},\text{spin}\}
\\
\end{tabularx}


\end{document}